\documentclass[11pt]{article}
\usepackage[varqu]{inconsolata}
\usepackage{subcaption}

\usepackage[preprint]{acl}

\usepackage{times}
\usepackage{latexsym}
\usepackage{amsmath}
\usepackage{booktabs}
\usepackage{amssymb}
\usepackage{dsfont}
\usepackage{multirow}
\usepackage{tcolorbox}
\tcbuselibrary{skins,breakable}
\usepackage{booktabs}
\usepackage{dsfont}          
\usepackage{hyperref}
\tcbuselibrary{skins}
\usepackage{enumitem}
\usepackage[table,dvipsnames]{xcolor} 

\usepackage[T1]{fontenc}

\usepackage{fontspec}
\usepackage{titletoc}
\usepackage[english]{babel}
 
\babelprovide[import]{bengali}
\babelprovide[import]{tamil}
\babelprovide[import]{hindi}      
\babelprovide[import]{gujarati}
 
\babelfont[bengali] {rm}[BoldFont=NotoSansBengali-Regular.ttf]   {NotoSansBengali-Regular.ttf}
\babelfont[tamil]   {rm}[BoldFont=NotoSansTamil-Regular.ttf]     {NotoSansTamil-Regular.ttf}
\babelfont[hindi]   {rm}[BoldFont=NotoSansDevanagari-Regular.ttf]{NotoSansDevanagari-Regular.ttf}
\babelfont[gujarati]{rm}[BoldFont=NotoSansGujarati-Regular.ttf]  {NotoSansGujarati-Regular.ttf}

\usepackage{pifont}
\newcommand{\xmark}{\ding{55}}
\usepackage{bm}

\newcommand{\name}{\textsc{Indi-RomCoM}}
\newcommand{\nameS}{\textsc{Indi-RomCoM} }

\tcbset{
  promptbox/.style={
    enhanced, breakable,
    colback=gray!6, colframe=gray!40,
    colbacktitle=gray!25,            
    coltitle=black,                  
    fonttitle=\bfseries\small,
    left=4pt, right=4pt, top=3pt, bottom=3pt,
    boxrule=0.4pt, arc=2pt,
  }
}

\usepackage{microtype}

\usepackage{graphicx}
\usepackage{float}

\title{Indi-RomCoM: Code-Mixed Benchmark for Evaluating LLMs on Romanized Indic-English Instructions}

\author{
 \textbf{Avisha Das\textsuperscript{1}},
 \textbf{Mihir Parmar\textsuperscript{2}},
 \textbf{Mohana Ramnath\textsuperscript{1}}, 
 {\normalfont and}
 \textbf{Pulkit Verma\textsuperscript{3}}
\\
 \textsuperscript{1}Shiv Nadar University Chennai,
 \textsuperscript{2}Google Cloud AI Research,
 \textsuperscript{3}IIT Madras
 \\
   \texttt{avishadas@snuchennai.edu.in}, 
   \texttt{mihirparmar@google.com},\\
   \texttt{mohana23110098@snuchennai.edu.in}, 
   \texttt{pulkitv@cse.iitm.ac.in}
}

\begin{document}
\maketitle
\begin{abstract}
Romanized Code Mixing (RCM), where bilingual speakers fluidly blend local languages with English in Roman script, has emerged as the dominant form of communication across multilingual communities. While Large Language Models (LLMs) perform strongly on monolingual and native-script benchmarks, their ability to follow instructions and reason over RCM-based content remains largely unexplored. To this end, we introduce the {\nameS}benchmark for facilitating systematic evaluation on \textbf{Indi}c \textbf{Rom}anized \textbf{Co}de-\textbf{M}ixed instructions. Our benchmark spans seven instruction-following tasks, four widely spoken Indic languages, and three controlled code-mixing intensity levels. We extensively evaluate a suite of LLMs covering proprietary, open-weight, and Indic-focused models under zero- and few-shot settings. LLMs consistently underperform on RCM instructions, with performance degrading as code-mixing density increases. Furthermore, reasoning tasks suffer less degradation than detection tasks (e.g., Toxicity) because the generated explanations offer necessary context. We believe {\nameS}helps the community in developing inclusive multilingual systems.

\end{abstract}

\section{Introduction}~\label{sec:intro}

Ensuring that Large Language Models (LLMs) can follow instructions in everyday regional language styles is a critical requirement for global deployment~\cite{gupta2024code, deuchar2020code}. India is a clear example of this challenge because it contains a highly diverse landscape of 22 official languages spoken by over 1.4 billion people. In this multilingual environment, English has mixed with local tongues to create a natural, rule-governed phenomenon called Romanized Code-Mixing (RCM)~\cite{sengupta2024social, winata-etal-2023-decades}. Driven by character limits, typing by sound on smartphones, and conversational convenience, millions of bilingual users fluidly blend regional Indic words, English terms, and local grammar into a single sentence, using the Latin alphabet exclusively.

\begin{figure}
    \centering
    \includegraphics[width=\linewidth]{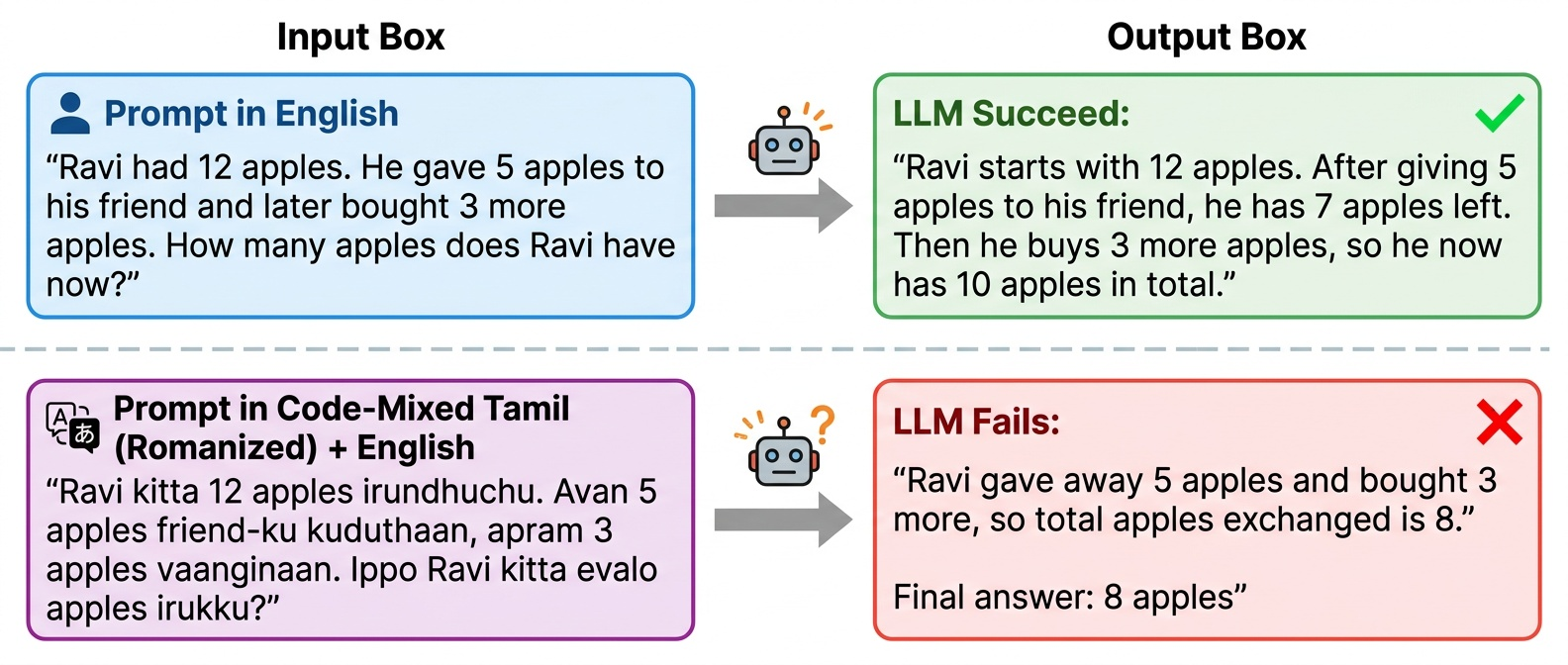}
    \caption{Example (LLaMa-7B-Instruct) showing failure to understand romanized “tanglish”.}
    \label{fig:teaser}
\end{figure}

For instance, a Hindi-English speaker asks, ``Yeh loan ke liye mujhe kya karna padega?'', which translates to, ``What do I have to do to get this loan?''. These sentences are highly structured everyday registers that dominate South Asian digital text. Yet, while generative AI models perform well on standard, single-language test sets, their accuracy drops sharply when processing these mixed linguistic setups, as exemplified in Figure \ref{fig:teaser} (Tamil-English). Open-weight LLMs built specifically for the Indian subcontinent, such as Airavata~\citep{gala2024airavata}, TamilLLaMA~\cite{balachandran2023tamil}, Sarvam-1~\citep{sarvam1_2024}, improve performance on native scripts, but their capabilities fail when confronted with Romanized text.

A primary obstacle is that the NLP community lacks the benchmarking systems needed to systematically diagnose these failure modes. Traditional code-mixed data collections like GLUECoS~\citep{khanuja2020gluecos} are limited to low-level, word-by-word classification tasks such as labeling parts of speech, making them unable to evaluate high-level reasoning. Conversely, newer generative benchmarks like CodeMixBench~\citep{yang-chai-2025-codemixbench} are limited only to native scripts. This effectively blinds them to the spelling and phonetic shifts that characterize human-model interactions. Refer to Table \ref{tab:related_work_comparison} for comparison across different state-of-the-art code-mixing benchmarks.

To bridge this evaluative gap, we introduce \textsc{Indi-RomCoM}, a systematic benchmark designed to stress-test large language models on Romanized Indic-English instructions. 
Covering four major regional languages across seven core task types, our framework utilizes a linguistically controlled generation pipeline to divide inputs into precise code-mixing intensity layers. 
Through this process, we construct a large-scale evaluation suite validated by native bilingual annotators to guarantee strong semantic preservation and contextual naturalness.
Using \nameS we evaluate 19 prominent language models under zero-shot and few-shot setups. 
Our experiments show that climbing code-mixing intensity drives severe, monotonic performance degradation across both general-purpose and specialized Indic open-weight architectures alike. 
A comprehensive structural breakdown of the framework is provided in Section \ref{sec:benchmark_full}.

In summary, we introduce \textbf{\name}, a multi-task benchmark developed using novel controlled \textbf{word-selection framework} to evaluate LLMs on RCM. Additionally, we present a comprehensive \textbf{empirical analysis} of widely used LLMs and introduce the \textbf{Register Defection Rate (RDR)} to quantify language switching failures during reasoning.

\section{Prior Work}

\begin{table}[t]
\centering
\vspace{-2mm}
\resizebox{\columnwidth}{!}{
\begin{tabular}{lcccc}
\toprule
\textbf{Work} & \textbf{Script} & \textbf{Languages} & \textbf{Task Type} & \textbf{Density} \\
\midrule
GLUECoS & NS & Global & Classification & \xmark \\
LinCE & NS & Global & Classification & \xmark \\
CodeMixBench & NS & Global & Generative & \xmark \\
ChiEngMixBench & NS & ZH,EN & Dialogue & \xmark \\
\citeauthor{choudhary-etal-2026-llms} & RS & HI,EN & Syntax & \xmark \\
IndicDB & NS,RS & Indic & Text to SQL & \xmark \\
IndicLLMSuite & NS & Indic & Resource & \xmark \\
UPDESH & NS & Indic & Instruction & \xmark \\
RomanSetu & RS & Indic,EN & Translation & \xmark \\
\midrule
\multirow{3}{*}{\textbf{\name}} & \multirow{3}{*}{\textbf{NS, RS}} & \multirow{3}{*}{\textbf{\shortstack{Indic,\\EN}}} & \textbf{Classification,} & \multirow{3}{*}{\textbf{\checkmark}} \\
 & & & \textbf{Generative,} & \\
 & & & \textbf{Instruction} & \\
\bottomrule
\end{tabular}%
}
\vspace{-2mm}
\caption{A comparison against existing resources. Abbreviations mean NS for Native Script, RS for Romanized Native Script, and Clf. for Classification. The Density column tracks if a benchmark evaluates performance across varying language mixing percentages. Our framework uniquely supports separate multi-density layers representing text sets with 25 percent and 50 percent and 75 percent word switching intensity.}
\vspace{-3mm}
\label{tab:related_work_comparison}
\end{table}

\begin{figure*}
    \centering
    \includegraphics[width=\textwidth]{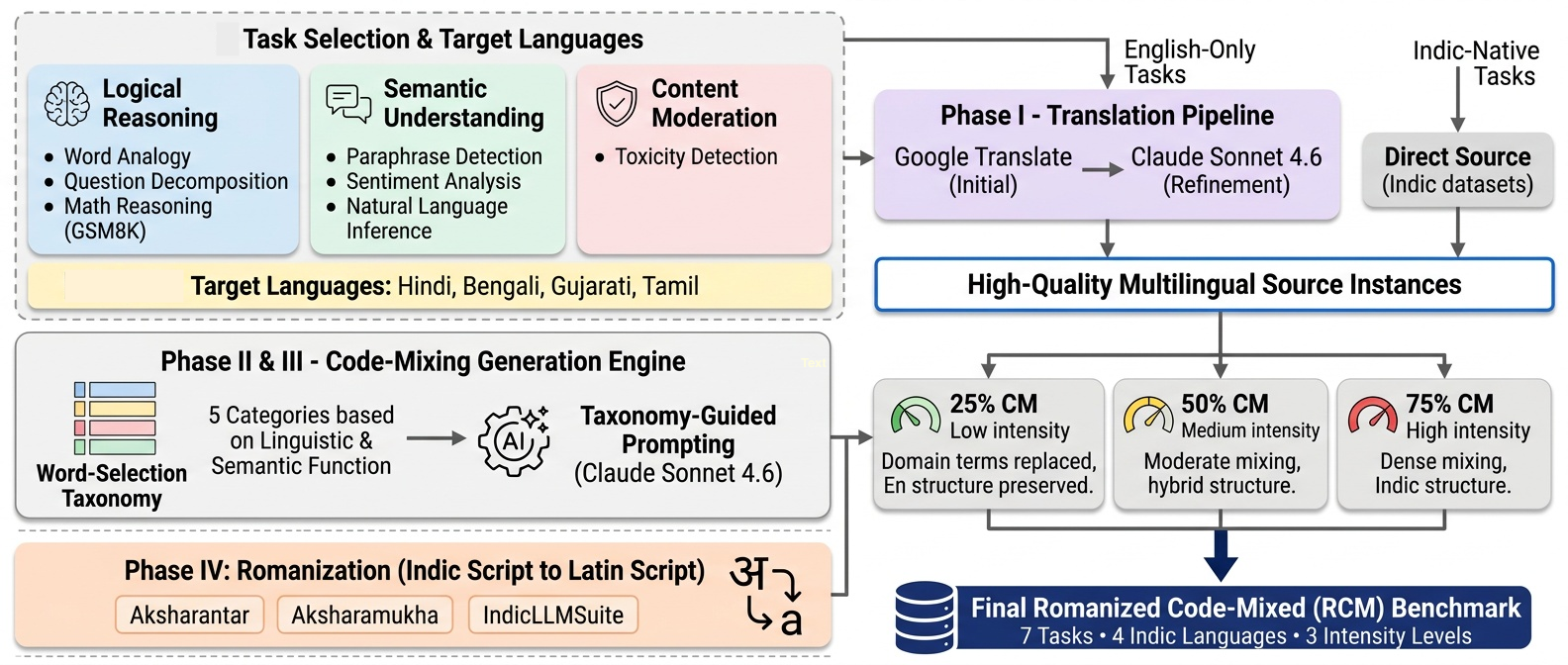}
    \caption{Overview of the \nameS creation framework. The pipeline processes English-only and Indic-native tasks through initial translation (Phase I), applies a taxonomy-guided generation engine to create controlled code-mixing at 25\%, 50\%, and 75\% intensities (Phases II \& III), and concludes with script romanization to produce the final multi-level benchmark (Phase IV).}
    \label{fig:pipeline}
\end{figure*}

\paragraph{Benchmarks for code-mixed NLP} 
Early computational modeling of code-mixed (CM) discourse focused primarily on token-level classification tasks or synthetic text generation. Frameworks like the GCM Toolkit~\citep{rizvi-etal-2021-gcm} built early foundations by computationally operationalizing structural linguistic constraints, such as the Equivalence Constraint and Matrix Language Frame theories, to synthesize data from parallel corpora. Similarly, \citet{sravani-mamidi-2023-enhancing} leveraged neural machine translation (NMT) architectures, employing filters like the Code-Mixing Index (CMI) and M-index to refine synthetic outputs. While these approaches advanced localized language modeling and individual datasets like GLUECoS~\citep{khanuja-etal-2020-gluecos} or LinCE~\citep{aguilar-etal-2020-lince} , they heavily relied on formal native scripts, targeted classification pipelines, and lacked the framework to explore multi-task instruction following.

As detailed in Table \ref{tab:related_work_comparison}, recent generative evaluations such as CodeMixBench~\citep{yang-chai-2025-codemixbench} and ChiEngMixBench~\citep{yang2026chiengmixbench} highlight significant LLM degradation on CM inputs, particularly in mathematical reasoning and dialogue.

\paragraph{Indic LLMs and Resources}
The development of Indic-focused models and datasets has accelerated to address South Asia's linguistic diversity. Large-scale resource efforts such as IndicLLMSuite~\citep{khan-etal-2024-indicllmsuite} and UPDESH~\citep{chitale2025updesh} provide extensive pretraining corpora across 13--22 languages, while specialized open-weight architectures including Airavata~\citep{gala2024airavata}, Tamil-Llama~\citep{balachandran2023tamil}, and Sarvam-1~\citep{sarvam1_2024} maximize localized syntactic representation.

Despite this progress, critical gaps persists. First, these suites remain anchored to native scripts, leaving CM-ed validation absent from official evaluation loops. Transliterated or CM-ed prompts collected through crowdsourcing (e.g., Airavata's Anudesh\footnote{\url{https://github.com/AI4Bharat/Anudesh}} pipeline) are rarely isolated for formal behavioral verification or benchmarked against varying density thresholds. Recent evaluations reinforce this urgency: IndicDB~\citep{dawar2026indicdb} exposed sharp execution drops under Romanized Hinglish instructions, and IndicSafe~\citep{pattnayak2026indicsafe}, despite auditing models for South Asian safety, overlooks the morphological blending inherent to real-world bilingual typing. The few localized studies that do address Romanization remain narrow in scope, e.g.,~\citet{choudhary-etal-2026-llms} restrict their analysis to verb-level inflections within Hindi-English pairs. 

{\name} addresses these gaps, providing the first human-validated, multi-task, multi-tier benchmark for Romanized code-mixed instruction-following, evaluating a comprehensive suite of LLMs (closed- and open-weights along with Indic) against the natural linguistic drift in informal Indic writing.

\section{\nameS Benchmark}
\label{sec:benchmark_full}
Figure~\ref{fig:pipeline} demonstrated all the phases of the \name benchmark. 
\subsection{Task Design and Selection}\label{sec:dataset}
We evaluate LLM performance on Romanized code-mixed instructions across seven
tasks spanning three competency dimensions: \textbf{logical reasoning},
\textbf{semantic understanding}, and \textbf{content moderation}. All tasks
are drawn from well-validated public sources with balanced label distributions,
ensuring that observed performance differences reflect genuine code-mixing
sensitivity rather than data artefacts~\cite{bai-etal-2024-longbench,
kwan-etal-2024-m4le}. For logical reasoning, we include Word Analogy and
Question Decomposition from Super-NaturalInstructions~\cite{wang2022super} and
Mathematical Reasoning from GSM8K~\cite{cobbe2021gsm8k}. For semantic
understanding, we draw on Paraphrase Detection, Sentiment Analysis, and Natural
Language Inference from the AI4Bharat IndicXtreme
collection~\cite{kunchukuttan2020ai4bharat, aggarwal2022indicxnli}. Toxicity
Detection is adapted from the IndicAlign-Toxic Matrix~\cite{chitale2025updesh}.
Table~\ref{tab:tasks} summarises the full task inventory; detailed descriptions
and dataset statistics are in Appendix~\ref{app:task_details}.
\paragraph{Languages} We evaluate romanized Indic-English code-mixed settings across four major Indian languages: Hindi, Bengali, Gujarati, and Tamil, covering both Indo-Aryan and Dravidian language families. These languages capture diverse linguistic structures, resource levels, and code-mixing patterns commonly observed in bilingual Indic-English communication~\cite{chitale2025updesh, jaavid2024romansetu}.

\begin{table*}
\centering
\resizebox{0.9\textwidth}{!}{%
\begin{tabular}{clll}
\toprule
\textbf{Type} & \textbf{Task} & \textbf{Source} & \textbf{Type} \\
\midrule
\multirow{3}{*}{\rotatebox[origin=c]{90}{\textbf{GEN}}}
 & Word Analogy     & SuperNI T1152-1159 \cite{wang2022super}           & Detection + Explanation \\  
 & Question Decomp.  & SuperNI T168/176/184 \cite{wang2022super}         & Explanation        \\ 
 & Math Reasoning   & GSM8K \cite{cobbe2021gsm8k}                       & Numerics + Explanation      \\ \midrule
\multirow{4}{*}{\rotatebox[origin=c]{90}{\textbf{CLF}}}
 & Paraphrase       & IndicXParaphrase \cite{kunchukuttan2020ai4bharat}  &  Detection        \\ 
 & Sentiment        & IndicSentiment \cite{kunchukuttan2020ai4bharat}   & Detection        \\ 
 & Inference        & IndicXNLI \cite{aggarwal2022indicxnli}            & Detection           \\ 
 & Toxicity         & IndicAlign-Toxic \cite{chitale2025updesh}         & Detection + Refusal     \\ \bottomrule
\end{tabular}%
}
\caption{\nameS task inventory. All instances are drawn from validated sources with stratified label distributions. Task type: GEN = Generation and CLF = Classification.}
\label{tab:tasks}
\end{table*}

\subsection{Code Mixing Framework}
We develop a four-step framework for generating controlled Indic-English RCM-ed instanced for the \name benchmark. Building upon the Equivalence Constraint~\cite{poplack2001code} and Matrix Language Frame~\cite{myers1993social} theories of bilingual switching, we systematically vary the mixing intensity while preserving semantic consistency. Full implementation details and prompting strategies, etc.
are in Appendix~\ref{app:cm_framework}. 

\paragraph{English to Indic Translation.} English-only tasks (Word Analogy, Question Decomposition~\cite{wang2022super}, and GSM8K~\cite{cobbe2021gsm8k}) are translated into indic versions following a two-step pipeline of machine translation and refinement. Other tasks are derived directly from their existing Indic datasets, IndicXNLI~\cite{aggarwal2022indicxnli}, IndicAlign-Toxic~\cite{chitale2025updesh}, IndicXParaphrase~\cite{khan2024indicllmsuite}, and IndicSentiment~\cite{gala2024airavata}. 

\paragraph{Word-Selection Taxonomy.} Each switchable word or phrase is assigned to one of five linguistically motivated categories~\cite{poplack2001code, winata-etal-2023-decades}, logged with its category label, Indic replacement, and a linguistic rationale to ensure substitutions reflect natural bilingual behavior rather than arbitrary replacement. Table~\ref{tab:cm_taxonomy} in Appendix summarizes the five categories, their switching policies, and illustrative cross-lingual examples.

\paragraph{Code-Mixing Intensity Levels.} Three intensity levels (25\%, 50\%, 75\%) are produced by progressively activating category subsets, with the Code-Mixing Index (CMI)~\cite{srivastava-singh-2021-challenges} quantifying the proportion of replaced English tokens. 

\paragraph{Romanization of Indic Script.} All Indic-script outputs are converted to Latin script using Aksharantar~\cite{madhani2023aksharantar} and Aksharamukha, with end-to-end romanizations sourced from IndicLLMSuite~\cite{khan2024indicllmsuite} where available. Natural transliteration variability is intentionally preserved to reflect real-world romanized code-mixed usage by multilingual South Asian users. 

\subsection{Benchmark Validation and Statistical Analysis}\label{sec:human_annotation}

\paragraph{Dataset Size} We sample 100 instances from each of seven tasks using stratified sampling over labels and difficulty levels, yielding 700 English base instances. Each instance is instantiated across the four Indic languages and three code-mixing intensity levels, resulting in \textbf{8,400 evaluation instances}. Task-wise sampling details are provided in Appendix~\ref{app:annotation}.

\paragraph{Human Validation} To assess the quality of \name, we conduct a human evaluation study. We recruit four native bilingual annotators (Tamil, Bengali, Hindi, Gujarati) with experience using Romanized code-mixed communication. Annotators independently evaluate samples across all tasks and CM levels using a 3-point Likert scale (\textit{1 = Unnatural, 2 = Somewhat Natural, 3 = Natural}), rating the code-mixed instances on two aspects: \textit{(i) Translation Fidelity}, measuring whether romanized code-mixed text preserves the semantic meaning of the English source using corpus-level Character n-gram F-score (ChrF;~\citealt{chitale2025updesh}), averaged across all seven tasks for each language. And, \textit{(ii) Code-mixing Naturalness}, assessing whether the code-mixed instances resembles bilingual usage rather than mechanical substitution. 

Two annotators  per language rate the same instances independently; we report Cohen's $\kappa$~\cite{cohen2013statistical} for pairwise naturalness agreement. Table~\ref{tab:iaa} summarizes per-language results. An average ChrF of \textbf{67.00} confirms substantial semantic overlap consistent with 50\%-CM intensity. Bengali and Tamil score highest on ChrF, reflecting denser preserved English structure, while Hindi and Gujarati score lower due to heavier Indic vocabulary switching. Overall agreement is moderate (average $\kappa = 0.615$) for code-mixing naturalness, consistent with the inherent subjectivity of naturalness assessment for code-mixed text~\cite{chitale2025updesh}. Agreement is highest for Gujarati ($0.905$) and lower for Bengali and Tamil due to stylistic variation in generation-heavy tasks. Higher code-mixing intensities generally produce lower agreement, especially for Question Decomposition, where dense predicate-level switching creates less common bilingual constructions. 
Per-task, per-language, per-CM-level results breakdown are provided in Appendix~\ref{app:annotation}.

\begin{table}
\centering
\begin{tabular}{lcc}
\toprule
\textbf{Language} & \textbf{ChrF} & \textbf{Cohen's $\kappa$} \\
 & \textbf{(Fidelity $\uparrow$)} & \textbf{(Naturalness $\uparrow$)} \\
\midrule
    Tamil      &   71.99 &       0.52 \\
    Bengali    &   74.08 &       0.48  \\
    Hindi      &   62.00 &       0.56 \\
    Gujarati   &   59.95  &       0.91 \\
\midrule
    \textbf{Average} & \textbf{67.00} & \textbf{0.615} \\
\bottomrule
\end{tabular}%
\caption{ChrF scores are reported between 50\%-CM romanized output-EN source.}
\label{tab:iaa}
\end{table}

\begin{figure}[t]
    \centering
    \includegraphics[width=\columnwidth]{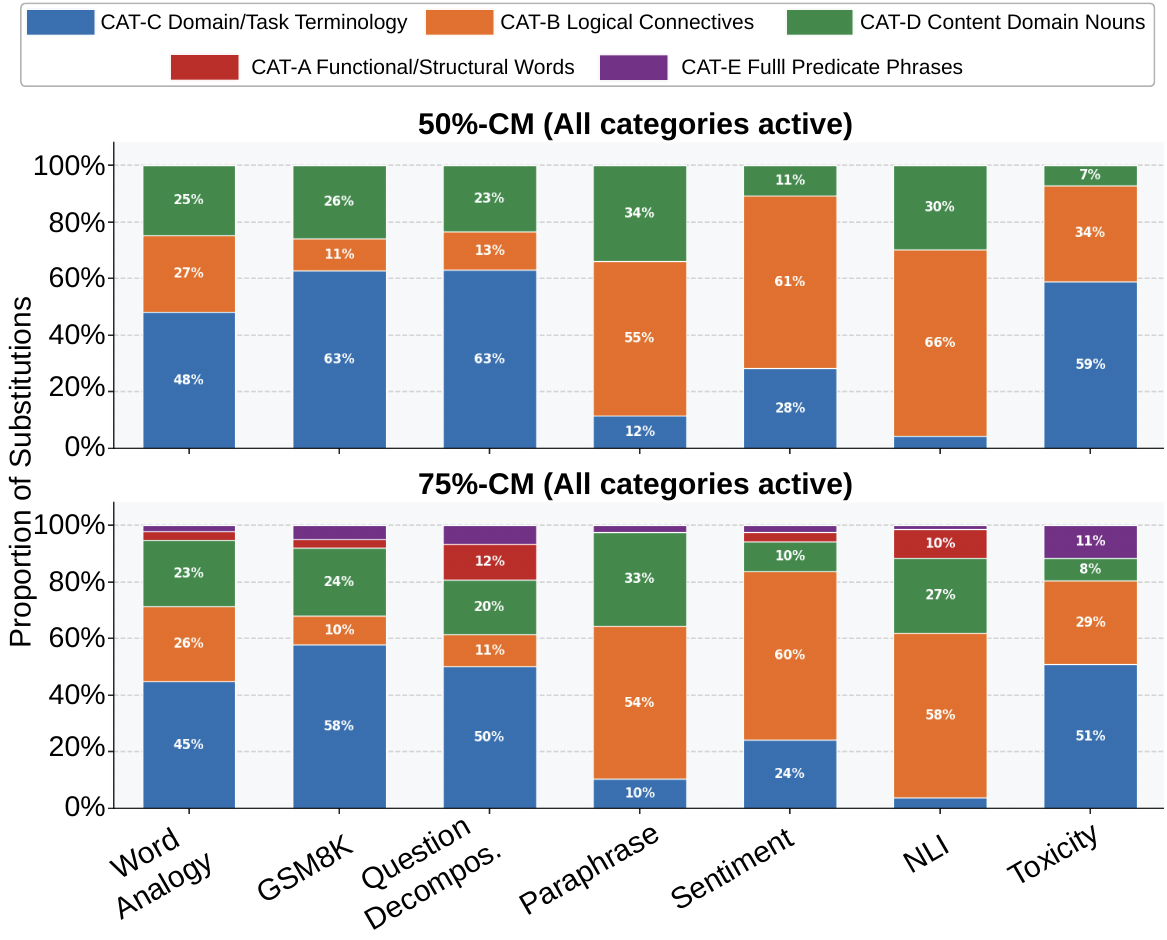}
    \caption{Word-category substitution frequency at 50\%-CM (top) and
    75\%-CM (bottom). At 25\%-CM all substitutions are CAT-C (100\%) by
    construction and are not shown. Colours: CAT-C (blue), CAT-B (orange),
    CAT-D (green), CAT-A (red), CAT-E (purple). Labels for
    segments $>$7\%.}
    \label{fig:cat_freq}
\end{figure}

\paragraph{\nameS Content Analysis}~\label{sec:content_analysis}
We analyze \nameS to verify if the three code-mixing intensity levels are distinct, category activations follow the intended linguistic progression, and task-level variation is systematic and not artifactual. The \textit{code-mixing quality} for each intensity level is studied using code-mixing index (CMI) metric. Table~\ref{tab:cmi_per_task} reports the mean CMI ($\mu \pm \sigma$) per task and CM level across all 8,400 instances.
Mean CMI rises monotonically: $\mu=0.249$ (25\%-CM) to $0.294$ (50\%-CM) to $0.360$ (75\%-CM). All pairwise differences are statistically significant (Wilcoxon signed-rank, $p < 10^{-100}$, Bonferroni-corrected), confirming that the three conditions represent genuinely distinct code-mixing regimes. We also perform the \textit{category-based frequency analysis} on the dataset's rationale annotations. 
Figures~\ref{fig:cat_freq}a--b show the proportional substitution breakdown at 50\%-CM and 75\%-CM respectively, where multi-category activation produces interpretable task-level variation. Additional details are reported in Appendix \ref{app:bm_content_analysis}.

 \begin{table}[t]
\centering
\resizebox{\columnwidth}{!}{%
\begin{tabular}{lccc}
\toprule
\textbf{Task} & \textbf{25\%} & \textbf{50\%} & \textbf{75\%} \\
\midrule
Word Analogy        & $0.423 \pm 0.252$ & $0.470 \pm 0.218$ & $0.503 \pm 0.190$ \\
GSM8K               & $0.084 \pm 0.147$ & $0.099 \pm 0.144$ & $0.152 \pm 0.175$ \\
Question Decomp.    & $0.179 \pm 0.147$ & $0.211 \pm 0.132$ & $0.365 \pm 0.123$ \\
Paraphrase          & $0.080 \pm 0.220$ & $0.186 \pm 0.254$ & $0.195 \pm 0.263$ \\
Sentiment           & $0.028 \pm 0.100$ & $0.078 \pm 0.115$ & $0.094 \pm 0.149$ \\
NLI                 & $0.004 \pm 0.016$ & $0.114 \pm 0.142$ & $0.138 \pm 0.157$ \\
Toxicity            & $0.851 \pm 0.171$ & $0.805 \pm 0.193$ & $0.981 \pm 0.049$ \\
\midrule
\textbf{All tasks}  & $0.249 \pm 0.338$ & $0.294 \pm 0.311$ & $0.360 \pm 0.345$ \\
\bottomrule
\end{tabular}}
\caption{Mean CMI ($\pm$ SD) per task and CM intensity level, averaged
across four languages and all instances per task.}
\label{tab:cmi_per_task}
\end{table}

\section{Experimental Setup}~\label{sec:experiments}

\subsection{Models} \label{sec:models}

We evaluate models across proprietary systems (OpenAI's GPT 3.5-turbo \cite{openai2026gpt35turbo}, Google's Gemini 3.5 Flash \cite{kavukcuoglu2026gemini35flash}, and Anthropic's Claude 4.6 Opus \cite{anthropic2026claudeopus46}), general open-weight architectures (Meta's LLaMa3 series \cite{grattafiori2024llama}, Alibaba Cloud's Qwen (2.5-3.5 series)~\cite{yang2025qwen3}, Google's Gemma4 series~\cite{team2024gemma}, and MistralAI's Mistral series \cite{mistralai2026modelsoverview}), Indic-focused multilingual and regional models, and specialized fine-tuned ablation variants to analyze performance across scale and training types. For our ablation study, we fine-tune LLaMA-3.2-3B, LLaMA-3.1-8B, and Gemma-2-7B on our training split using Low-Rank Adaptation to assess the necessity of explicit code-mixed supervision. A complete model list, fine-tuning configurations, comprehensive hyperparameter details, and dataset stratification metrics are provided in Appendix~\ref{app:models}.
 
\subsection{Prompting Protocol}
\label{sec:prompting}

\begin{table*}
\footnotesize
\centering
\setlength{\extrarowheight}{2pt}
\resizebox{\textwidth}{!}{%
\begin{tabular}{cp{9.2cm}p{5.8cm}}
\toprule
\textbf{Task} & \textbf{Template Format and Prompt Frame} & \textbf{Running Example (50\%-CM)} \\[2pt]
\midrule

\multirow{8}{*}{\rotatebox[origin=c]{90}{\textsc{\textcolor{purple}{\shortstack{Sentiment\\Analysis}}}}} &
\textcolor{orange!80!black}{\textbf{Task}} Classify the sentiment of the following product review as positive (1), neutral (0), or negative ($-$1). \newline
\textit{[Few-shot block omitted in zero-shot condition]} \newline
\textbf{Review} $\langle$D1$\rangle$ \quad \textbf{Sentiment} $\langle$L1$\rangle$ \quad $\vdots$ \newline
\textcolor{blue!70!black}{\textbf{[Query]}} \textbf{Review} $\langle$in$\rangle$ \quad \textbf{Sentiment} \newline
\textcolor{blue!70!black}{\textbf{Register}} Please respond in the same language and script as the input. &
\textcolor{teal}{\textbf{Tamil}} \newline
\textbf{Review} Foldable type of microphone with mic marrum micro sd card slot. \newline
\textbf{Sentiment} 1 \\[2pt]
\hline\\[-5pt]

\multirow{8}{*}{\rotatebox[origin=c]{90}{\textsc{\textcolor{purple}{\shortstack{Inference\\(NLI)}}}}} &
\textcolor{orange!80!black}{\textbf{Task}} Given a premise and a hypothesis, determine whether the hypothesis is entailed (1), neutral (0), or contradicted (2) by the premise. \newline
\textit{[Few-shot block omitted in zero-shot condition]} \newline
\textbf{Premise} $\langle$D1p$\rangle$ \quad \textbf{Hypothesis} $\langle$D1h$\rangle$ \quad \textbf{Label} $\langle$L1$\rangle$ \quad $\vdots$ \newline
\textcolor{blue!70!black}{\textbf{[Query]}} \textbf{Premise} $\langle$p$\rangle$ \quad \textbf{Hypothesis} $\langle$h$\rangle$ \quad \textbf{Label} \newline
\textcolor{blue!70!black}{\textbf{Register}} Please respond in the same language and script as the input. &
\textcolor{teal}{\textbf{Hindi}} \newline
\textbf{Premise} Waise, yeh the three U2 pilots jo President Kennedy ke saath Washington mein General May ke daftar mein the. \newline
\textbf{Hypothesis} General May aur pilots ne daftar mein achha waqt bitaya. \newline
\textbf{Label} 1 \\[2pt]
\hline\\[-5pt]

\multirow{8}{*}{\rotatebox[origin=c]{90}{\textsc{\textcolor{purple}{Paraphrase}}}} &
\textcolor{orange!80!black}{\textbf{Task}} Determine whether the two sentences are paraphrases of each other. Output 1 if they are paraphrases and 0 otherwise. \newline
\textit{[Few-shot block omitted in zero-shot condition]} \newline
\textbf{S1} $\langle$D1s1$\rangle$ \quad \textbf{S2} $\langle$D1s2$\rangle$ \quad \textbf{Label} $\langle$L1$\rangle$ \quad $\vdots$ \newline
\textcolor{blue!70!black}{\textbf{[Query]}} \textbf{S1} $\langle$s1$\rangle$ \quad \textbf{S2} $\langle$s2$\rangle$ \quad \textbf{Label} \newline
\textcolor{blue!70!black}{\textbf{Register}} Please respond in the same language and script as the input. &
\textcolor{teal}{\textbf{Bengali}} \newline
\textbf{S1} Police station-e giye ami complaint korlam je amar phone churi hoyeche. \newline
\textbf{S2} Ami police station-e giye janalam je amar phone ta churi giyeche. \newline
\textbf{Label} 1 \\[2pt]
\hline\\[-5pt]

\multirow{8}{*}{\rotatebox[origin=c]{90}{\textsc{\textcolor{purple}{Toxicity}}}} &
\textcolor{orange!80!black}{\textbf{Task}} Classify whether the following text is toxic or non-toxic. Then briefly explain your classification in the same register as the input. \newline
\textit{[Few-shot block omitted in zero-shot condition]} \newline
\textbf{Text} $\langle$D1$\rangle$ \quad \textbf{Label} $\langle$L1$\rangle$ \quad \textbf{Expl.} $\langle$E1$\rangle$ \quad $\vdots$ \newline
\textcolor{blue!70!black}{\textbf{[Query]}} \textbf{Text} $\langle$in$\rangle$ \quad \textbf{Label} \quad \textbf{Expl.} \newline
\textcolor{blue!70!black}{\textbf{Register}} Please respond in the same language and script as the input. &
\textcolor{teal}{\textbf{Hindi}} \newline
\textbf{Text} Tu bilkul bekaar hai, koi tumhari baat nahi sunta kyunki tum ek loser ho. \newline
\textbf{Label} toxic \newline
\textbf{Expl.} Is tarah ki baat karna galat hai kyunki yeh doosre insaan ki izzat ko deliberately hurt karta hai. \\[2pt]
\hline\\[-5pt]

\multirow{8}{*}{\rotatebox[origin=c]{90}{\textsc{\textcolor{purple}{Mathematics}}}} &
\textcolor{orange!80!black}{\textbf{Task}} Solve the following word problem step by step. Show all intermediate calculations using the format \texttt{<<expr=result>>}, then output the final numeric answer after \texttt{\#\#\#\#}. \newline
\textit{[Few-shot block omitted in zero-shot condition]} \newline
\textbf{Problem} $\langle$D1$\rangle$ \quad \textbf{Solution} $\langle$E1$\rangle$ \quad \textbf{Answer} $\langle$A1$\rangle$ \quad $\vdots$ \newline
\textcolor{blue!70!black}{\textbf{[Query]}} \textbf{Problem} $\langle$in$\rangle$ \quad \textbf{Solution} \newline
\textcolor{blue!70!black}{\textbf{Register}} Please respond in the same language and script as the input. &
\textcolor{teal}{\textbf{Hindi}} \newline
\textbf{Problem} At the bake sale, Tamara makes \$32 from brownies. She made 2 pans cut into 8 pieces each. How much did each brownie cost? \newline
\textbf{Solution} Tamara ke paas 2$\times$8$=$\texttt{<<16>>}16 tukde the. Har ek ki qeemat \$32/16$=$\$\texttt{<<2>>}2 thi. \newline
\textbf{Answer} 2 \\[2pt]
\hline\\[-5pt]

\multirow{8}{*}{\rotatebox[origin=c]{90}{\textsc{\textcolor{purple}{\shortstack{Question\\Decomposition}}}}} &
\textcolor{orange!80!black}{\textbf{Task}} Given a question, its answer, and supporting facts, decompose it into sub-questions each isolating one verifiable fact. Format as Step1~..., Step2~..., with the Wikipedia source page for each step. \newline
\textit{[Few-shot block omitted in zero-shot condition]} \newline
$\langle$D1$\rangle$ \quad $\vdots$ \newline
\textcolor{blue!70!black}{\textbf{[Query]}} $\langle$in$\rangle$ \quad \textbf{Decomposition} \newline
\textcolor{blue!70!black}{\textbf{Register}} Please respond in the same language and script as the input. &
\textcolor{teal}{\textbf{Tamil}} \newline
\textbf{Step1} Is Mr. Peanut alive? \textbf{Step2} Do you have to be alive to compose something? \textbf{Step3} Are answers to \#1 and \#2 the same? \newline
\textbf{Expl.} This is a strategic vivaribhbhu into 3 ubha-ghelvi(ghal). Each bhadhi isolates a jharibharghghaghghudhiya unmai to bhadhil the yes/no ghelvi. \\[2pt]
\hline\\[-5pt]

\multirow{8}{*}{\rotatebox[origin=c]{90}{\textsc{\textcolor{purple}{\shortstack{Word\\Analogy}}}}} &
\textcolor{orange!80!black}{\textbf{Task}} Complete the analogy. Given A~B.~C~?, find D such that C~D holds the same relationship as A~B. Output the answer word, then explain the relationship type. \newline
\textit{[Few-shot block omitted in zero-shot condition]} \newline
$\langle$D1$\rangle$ \quad $\vdots$ \newline
\textcolor{blue!70!black}{\textbf{[Query]}} $\langle$in$\rangle$ \quad \textbf{Answer} \newline
\textcolor{blue!70!black}{\textbf{Register}} Please respond in the same language and script as the input. &
\textcolor{teal}{\textbf{Gujarati}} \newline
\textbf{Input} throw fly. aspire~? \newline
\textbf{Answer} attain \newline
\textbf{Expl.} Fly ej throw no parinam che. Aaथी, aa uvamai pramane, attain ej aspire no parinam che. \\[2pt]

\bottomrule
\end{tabular}%
}
\caption{Structural breakdown of our instruction-following dataset tasks. \textcolor{orange!80!black}{\textbf{Orange}} and \textcolor{blue!70!black}{\textbf{blue}} labels denote task definition and query components respectively. \textcolor{teal}{\textbf{Teal}} tags indicate the example language: Tamil (Sentiment, Decomposition), Hindi (NLI, Toxicity, Mathematics), Bengali (Paraphrase), Gujarati (Word Analogy), all at 50\%-CM. Placeholders: $\langle$D$k\rangle$ = $k$-th demo input, $\langle$L$k\rangle$ = label, $\langle$E$k\rangle$ = explanation, $\langle$A$k\rangle$ = answer, $\langle$in$\rangle$ = query input.}
\label{tab:reframing_examples}
\end{table*}

We evaluate all models under zero-shot and stratified 3-shot settings using structured four-part prompt templates that enforce explicit register maintenance rules. The underlying evaluations span seven core instruction-following tasks where input values are dynamically mapped to specific romanized code-mixed intensity configurations. Complete structural prompt definitions, few-shot conditioning frameworks, and individual task-specific instruction boxes are also provided in Appendix~\ref{app:prompt_details}.

\subsection{Evaluation Metrics}
\label{sec:metrics}
We use four metrics to evaluate model performance across our benchmark. (i) \textbf{Task Accuracy (ACC)} measures exact string matches for classification alongside standard extraction rubrics for generation tasks; (ii)~\textbf{Register Defection Rate (RDR)} tracks the proportion of model outputs that incorrectly revert to standard English text; (iii)~\textbf{Vocabulary Coverage Rate (VCR)} measures the percentage of unique romanized tokens that exist as complete entries in a model tokenizer vocabulary; and (iv)~\textbf{Performance Gap (PG)} calculates the absolute accuracy loss between the English baseline and each code-mixed intensity level. We validate these trends using Wilcoxon signed-rank tests, Bonferroni corrections, and Spearman correlation checks. Detailed definitions, parsing criteria, and statistical configurations are provided in Appendix~\ref{app:metrics_details}.

\section{Results and Analysis}

\subsection{Main results}
\begin{figure*}[!htb]
    \centering
    \includegraphics[width=\textwidth]{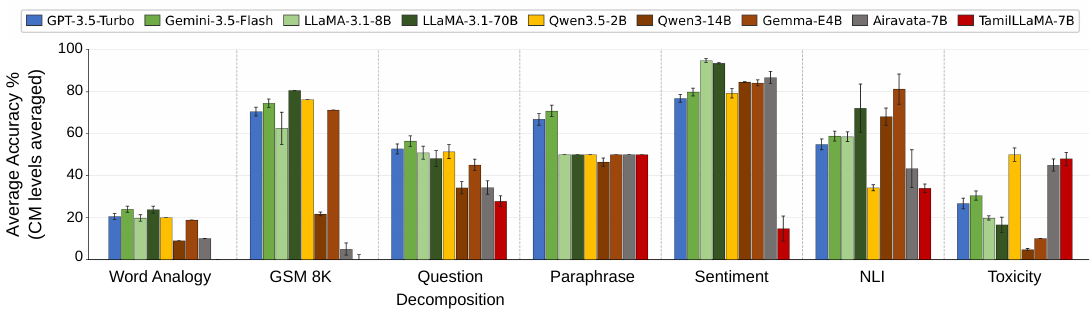}
    \caption{%
        \textbf{Average task accuracy (\%) under code-mixed (CM) instructions by model and task.}
        Each bar reports accuracy averaged across three CM intensity levels (25\%, 50\%, 75\%) 
        and four Indic languages in the zero-shot setting.
        Error bars denote standard deviation.
        Dashed vertical lines separate the seven task groups.
    }
    \label{fig:cm_accuracy_by_task}
\end{figure*}

We present the evaluation results on the \nameS benchmark. Table~\ref{tab:h1_main} reports average task accuracy across all seven tasks for the 19 models in comparison with the English baseline (BL) and three code-mixing intensity levels. Figure~\ref{fig:cm_accuracy_by_task} illustrates per-task performance variations across a selected set of best performing models and code-mixing settings. Additionally, Table~\ref{tab:rdr} reports the Register Defection Rate (RDR) across all models and CM conditions, measuring the proportion of responses that revert primarily to English despite receiving code-mixed instructions.

\begin{table*}[!htb]
\centering
\resizebox{\textwidth}{!}{%
\begin{tabular}{llcccccccc}
\toprule
\textbf{Family} & \textbf{Model}
  & \textbf{BL} & \textbf{25\%-CM} & \textbf{50\%-CM} & \textbf{75\%-CM}
  & \textbf{PG-25} & \textbf{PG-50} & \textbf{PG-75} \\
\midrule
\multirow{3}{*}{Closed}
  & GPT-3.5-Turbo     & $52.3 \pm 28.4$ & $46.1 \pm 25.2$ & $44.8 \pm 24.7$ & $44.3 \pm 24.5$ & $+6.2$ & $+7.5$ & $+8.0$ \\
  & Claude Opus 4.6   & $68.7 \pm 22.1$ & $63.4 \pm 21.8$ & $61.9 \pm 21.5$ & $61.2 \pm 21.3$ & $+5.3$ & $+6.8$ & $+7.5$ \\
  & Gemini 3.5 Flash  & $61.4 \pm 25.9$ & $56.2 \pm 24.3$ & $54.8 \pm 23.9$ & $54.1 \pm 23.7$ & $+5.2$ & $+6.6$ & $+7.3$ \\
\midrule
\multirow{3}{*}{LLaMA}
  & LLaMA-3.2-1B                        & $24.3 \pm 16.9$          & $11.5 \pm 4.5$           & $11.6 \pm 5.4$           & $11.6 \pm 5.3$           & $+12.8$ & $+12.7$ & $+12.8$ \\
    & LLaMA-3.1-8B                        & $50.1 \pm 33.6$          & $48.6 \pm 36.0$          & $47.8 \pm 35.3$          & $47.9 \pm 35.4$          & $+1.5$  & $+2.4$  & $+2.2$  \\
  & LLaMA-3.1-70B               & $54.6 \pm 32.3$ & $51.5 \pm 28.8$ & $50.6 \pm 28.3$ & $50.5 \pm 28.4$ & $+3.0$  & $+4.0$  & $+4.0$  \\

\midrule
\multirow{2}{*}{Mistral}
  & Mistral-3B & $37.2 \pm 12.7$ & $24.8 \pm 11.1$ & $22.1 \pm 18.3$ & $19.7 \pm 11.9$ & $+12.4$ & $+15.1$ & $+17.5$ \\
  & Mistral-7B & $49.7 \pm 35.7$          & $44.3 \pm 35.0$          & $43.6 \pm 34.5$          & $43.3 \pm 34.3$          & $+5.4$ & $+6.1$ & $+6.4$ \\
\midrule
\multirow{6}{*}{Qwen}
  & Qwen2.5-1.5B                  & $43.4 \pm 27.0$          & $40.4 \pm 28.0$          & $40.5 \pm 28.0$          & $40.4 \pm 28.0$          & $+3.1$ & $+2.9$ & $+3.0$ \\
  & Qwen2.5-7B                    & $49.7 \pm 35.7$          & $44.3 \pm 35.0$          & $43.6 \pm 34.5$          & $43.3 \pm 34.3$          & $+5.4$ & $+6.1$ & $+6.4$ \\
  & Qwen3-4B                      & $32.3 \pm 15.2$          & $34.4 \pm 23.4$          & $33.3 \pm 21.5$          & $32.9 \pm 21.3$          & $-2.1$ & $-1.0$ & $-0.6$ \\
  & Qwen3-14B                     & $29.5 \pm 24.8$          & $34.7 \pm 33.4$          & $34.0 \pm 32.4$          & $33.8 \pm 32.2$          & $-5.2$ & $-4.5$ & $-4.3$ \\
  & Qwen3.5-2B           & $58.0 \pm 40.2$ & $51.1 \pm 35.1$ & $51.6 \pm 36.1$ & $51.4 \pm 36.2$ & $+7.0$ & $+6.4$ & $+6.6$ \\
  & Qwen3.5-9B                    & $49.7 \pm 35.7$          & $44.3 \pm 35.0$          & $43.6 \pm 34.5$          & $43.3 \pm 34.3$          & $+5.4$ & $+6.1$ & $+6.4$ \\
\midrule
\multirow{2}{*}{Gemma}
  & Gemma-E2B          & $43.0 \pm 30.9$ & $42.0 \pm 24.2$ & $40.9 \pm 23.3$ & $40.7 \pm 23.2$ & $+1.1$ & $+2.1$ & $+2.3$ \\
  & Gemma-E4B & $48.0 \pm 37.8$ & $46.1 \pm 36.7$ & $44.6 \pm 34.9$ & $44.5 \pm 34.7$ & $+2.0$ & $+3.4$ & $+3.6$ \\
\midrule
\multirow{3}{*}{Indic}
  & Airavata-7B   & $38.6 \pm 36.1$ & $36.4 \pm 32.9$ & $33.6 \pm 30.5$ & $32.9 \pm 30.4$ & $+2.1$ & $+5.0$ & $+5.7$ \\
  & TamilLLaMA-7B & $26.9 \pm 36.8$ & $28.6 \pm 35.6$ & $27.4 \pm 35.8$ & $27.4 \pm 35.9$ & $-1.7$ & $-0.6$ & $-0.6$ \\
  & Sarvam-30B    & $64.2 \pm 24.5$ & $59.8 \pm 23.1$ & $57.3 \pm 22.8$ & $56.1 \pm 22.4$ & $+4.4$ & $+6.9$ & $+8.1$ \\
\bottomrule
\end{tabular}%
}
\caption{%
  Average task accuracy (\%) across 7 tasks in zero-shot setting, averaged over 4 languages ($\pm$ SD across tasks).
  BL = English baseline. PG = Performance Gap (BL $-$ CM level; positive = accuracy drop).
  CM = Code Mixing. \textbf{Bold} = best BL per family.
}
\label{tab:h1_main}
\end{table*}

\begin{table*}[!htb]
\centering
\resizebox{\textwidth}{!}{%
\begin{tabular}{llcccc}
\toprule
\textbf{Family} & \textbf{Model}
  & \textbf{RDR @ 25\%-CM} & \textbf{RDR @ 50\%-CM} & \textbf{RDR @ 75\%-CM}
  & \textbf{Avg RDR} \\
\midrule
\multirow{3}{*}{Closed}
  & GPT-3.5-Turbo    & $48.3$ & $53.7$ & $61.2$ & $54.4$ \\
  & Claude Opus 4.6  & $31.4$ & $35.8$ & $40.6$ & $35.9$ \\
  & Gemini 3.5 Flash & $39.2$ & $44.1$ & $50.3$ & $44.5$ \\
\midrule
\multirow{3}{*}{LLaMA}
  & LLaMA-3.2-1B     & $51.0$ & $49.6$ & $49.4$ & $50.0$ \\
  & LLaMA-3.1-8B     & $99.8$ & $99.7$ & $99.7$ & $99.7$ \\
  & LLaMA-3.1-70B    & $97.6$ & $97.5$ & $97.5$ & $97.5$ \\
\midrule
\multirow{2}{*}{Mistral}
  & Mistral-3B       & $77.8$ & $89.3$ & $91.7$ & $86.3$ \\
  & Mistral-7B       & $60.6$ & $64.7$ & $65.1$ & $63.5$ \\
\midrule
\multirow{6}{*}{Qwen}
  & Qwen2.5-1.5B     & $99.8$ & $99.8$ & $99.8$ & $99.8$ \\
  & Qwen2.5-7B       & $60.6$ & $64.7$ & $65.1$ & $63.5$ \\
  & Qwen3-4B         & $99.9$ & $98.7$ & $97.8$ & $98.8$ \\
  & Qwen3-14B        & $89.7$ & $91.3$ & $95.1$ & $92.0$ \\
  & Qwen3.5-2B       & $99.6$ & $99.6$ & $99.6$ & $99.6$ \\
  & Qwen3.5-9B       & $60.6$ & $64.7$ & $65.1$ & $63.5$ \\
\midrule
\multirow{2}{*}{Gemma}
  & Gemma-E2B        & $78.3$ & $82.7$ & $93.5$ & $84.8$ \\
  & Gemma-E4B        & $74.8$ & $81.3$ & $84.3$ & $80.1$ \\
\midrule
\multirow{3}{*}{Indic}
  & Airavata-7B      & $10.4$ & $18.4$ & $20.7$ & $16.5$ \\
  & TamilLLaMA-7B    & $16.6$ & $21.7$ & $21.3$ & $19.9$ \\
  & Sarvam-30B       & $22.4$ & $26.8$ & $29.1$ & $26.1$ \\
\bottomrule
\end{tabular}%
}
\caption{%
  Register Defection Rate (RDR, \%) when the query is in CM register. Lower RDR indicates stronger register adherence.  Avg RDR is the mean across the three CM intensity levels.
}
\label{tab:rdr}
\end{table*}

\textbf{Poor performance of models on Indic RCM-ed data as compared to English}~\label{sec:h1} Table~\ref{tab:h1_main} reveals a consistent pattern of accuracy degradation under Romanized code-mixed (RCM) conditions across nearly all models. Every model except the Qwen3 variants records a positive performance gap, proving Indic RCM input is substantially harder than English for current LLMs. Even the strongest models are affected: Claude Opus 4.6 drops from 68.7\% at baseline to 61.2\% at 75\%-CM (PG-75 $= +7.5$), and Sarvam-30B, despite dedicated Indic pretraining (falls from 64.2\% to 56.1\% (PG-75 $= +8.1$)). Smaller models suffer steeper drops; Mistral-3B loses 17.5 percentage points by 75\%-CM. These results demonstrate that no model family is robust to Indic RCM input, exposing a critical gap in multilingual LLM evaluation.

\textbf{Model performance varies by task type}~\label{sec:h2} 
Figure~\ref{fig:cm_accuracy_by_task} reveals task-level variation in model robustness under RCM input. GSM8K yields the highest accuracy overall (up to $\sim$80\% for LLaMA-3.1-70B), consistent with CAT-0 protection of mathematical tokens insulating problem content from CM disruption. Sentiment analysis achieves the most consistent cross-model performance, suggesting that naturally code-mixed product-review vocabulary aids rather than hinders comprehension.
Conversely, Word Analogy and Toxicity detection show the lowest absolute accuracy across all models, the former due to morphological sensitivity, the latter exposing a critical safety gap: models trained on English safety data struggle to detect harm expressed in CM register, with most models scoring below 50\%, a directly consequential failure for real-world deployment.

\textbf{Indic Pretraining Drives Register Adherence} Table~\ref{tab:rdr} shows that most open-source models defect overwhelmingly into English under
RCM input: LLaMA-3.1-8B, Qwen2.5-1.5B, and Qwen3-4B all exceed 98\% Avg RDR, regardless of
scale. The only models that genuinely maintain CM register are Indic-trained: Airavata-7B
(16.5\%), TamilLLaMA-7B (19.9\%), and Sarvam-30B (26.1\%). Notably, RDR rises with CM intensity
across nearly all models, confirming that higher mixing pressure amplifies register defection
rather than suppressing it.

\subsection{Ablation Studies}

\textbf{Few-Shot Prompting help close the performance gap} Three-shot CM demonstrations yield a consistent 7--8\% accuracy gain over zero-shot across all model families (Appendix Table~\ref{tab:h2_fewshot}), driven primarily by a sharp reduction in register defection on classification tasks, providing in-register examples anchors the output format. However, performance gaps (PG) remain largely unchanged, confirming that CM examples help models adopt the correct register without improving their underlying comprehension of code-mixed content, and gains on GSM8K are minimal, replicating prior findings on mathematical reasoning~\cite{yang2025codemixbench}. The results are shown in Appendix Section~\ref{app:results}.

\begin{figure*}[!htb]
    \centering
    \includegraphics[width=\textwidth]{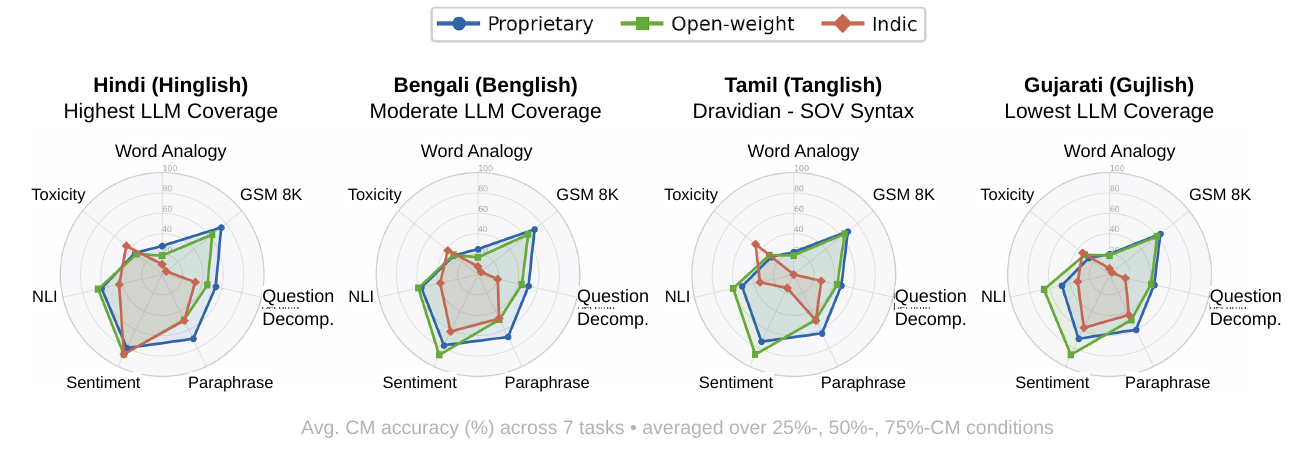}
    \caption{%
        \textbf{Per-language CM accuracy across 7 tasks and 3 LLM families.}
        Each spider plot shows average accuracy (\%) across 25\%-, 50\%-, and 75\%-CM conditions
        for Proprietary (\textcolor[HTML]{2166AC}{blue}),
        Open-weight (\textcolor[HTML]{4DAC26}{green}),
        and Indic-focused (\textcolor[HTML]{D6604D}{red}) model families.
        Performance decreases from Hindi to Gujarati, reflecting differences in
        LLM pretraining corpus coverage across languages.
    }
    \label{fig:spider_languages}
\end{figure*}

\textbf{Model performance varies across languages}
Figure~\ref{fig:spider_languages} shows that Hindi (Hinglish) achieves the highest average CM accuracy across all three model families, consistent with its greater representation in LLM pretraining corpora~\cite{choudhary-etal-2026-llms}. Bengali (Banglish) performs moderately, while Tamil (Tanglish) exhibits sharper degradation on generation tasks, attributable to its Dravidian SOV syntax producing more complex CM structures. Gujarati (Gujlish) is the most challenging language overall, reflecting its minimal CommonCrawl representation and near-zero coverage in existing Indic-focused pretraining.

\textbf{Finetuning ablation} To assess whether targeted supervision can bridge the code-mixing gap, we fine-tuned LLaMA-3.2-3B and LLaMA-3.1-8B via QLoRA (rank\,=\,16, $\alpha$\,=\,32, 4-bit NF4 quantisation) on 50\%-CM Romanised instruction pairs spanning all four languages and seven tasks.
Both models converge to near-identical aggregate exact-match rates ($86.3\%$ vs.\ $86.4\%$), confirming that scale yields diminishing returns once the CM format is supervised directly, while the near-total register-defection rate ($99.9\%$) reflects the Roman-script nature of the outputs rather than generation failure. Detailed results are presented in Appendix Table~\ref{tab:finetuning_ablation}.

\section*{Conclusion}
We presented \textsc{Indi-RomCoM}, a benchmark to stress-test large language models on Romanized Indic-English instructions across multiple languages, tasks, and intensity levels. 
Our extensive evaluation of 17 prominent language models reveals that climbing code-mixing density drives severe, monotonic performance degradation across both general-purpose and specialized architectures. 
Ablation variants show that this systematic challenge cannot be resolved by model scale or native-script pretraining alone, proving that targeted code-mixed supervision is necessary. 
This framework establishes a rigorous foundation for future research into register-robust multilingual models that reflect naturalistic bilingual communication.

\section*{Limitations}
\nameS has several limitations worth acknowledging. The controlled generation pipeline may underrepresent organic real-world code-mixing phenomena such as phonetic drift and intra-word mixing. Human validation relied on only four annotators, which is too small a pool to capture sociodemographic variation in code-mixing norms. Coverage is restricted to four Indic languages, leaving out major Indic language communities such as Telugu, Kannada, and Marathi. Additionally, the benchmark inherits labeling noise from the upstream crowdsourced datasets it draws from, and several instruction-following categories including summarization and open-ended question answering are absent from the current task suite.

\section*{Ethical Considerations}
All tasks are derived from publicly available datasets, and no user data was collected from private communications or social media. The Toxicity Detection task contains genuinely harmful content by design because evaluating model refusal behavior under code-mixed framing is an important safety challenge. However, the benchmark should not be used to generate or amplify harmful content. Annotators rating toxic instances were informed in advance and could skip individual samples at any time. We also caution against benchmark overfitting, where models are tuned to specific intensity levels without genuinely improving on naturalistic code-mixed understanding.

\bibliography{custom}

\cleardoublepage
\appendix

\titlecontents{section*}[1.5em]
  {\vspace{0.5em}\bfseries\color{blue}}
  {\contentslabel{1.5em}}
  {\hspace*{-1.5em}}
  {\titlerule*[0.5pc]{.}\contentspage}

\titlecontents{subsection}[3.8em]
  {\normalfont\color{blue}}
  {\contentslabel{2.3em}}
  {\hspace*{-2.3em}}
  {\titlerule*[0.5pc]{.}\contentspage}

\twocolumn[{
  \vspace{2em}
  \begin{center} 
    \Large\textbf{\textsf{Table of Contents for Appendix}}
  \end{center}
  \vspace{1em}

  \startcontents[appendix]
  \printcontents[appendix]{}{1}{\setcounter{tocdepth}{2}}
  \vspace{2em}
}]

\section{Task Details} ~\label{app:task_details}
\subsection{Data Description}
Table~\ref{tab:tasks-counts-appendix} reports the original number of instances drawn from each source dataset before any preprocessing, filtering, or code-mixing. These counts represent the full, unmodified benchmark splits as released by their respective authors. All instances were retained to construct the code-mixed and romanised variants described in Section~\ref{sec:dataset}, yielding the training corpus used in our fine-tuning ablation experiments. 

\begin{table}[h]
\centering
\rowcolors{2}{gray!13}{}
\resizebox{\columnwidth}{!}{
\begin{tabular}{llr}
\toprule
\textbf{Task} & \textbf{Label Schema} & \textbf{N} \\
\midrule
Word Analogy     & Correct (1) or Incorrect (0) and Explanation  &     48 \\
Question Decomp. & Sub-Questions Decomposition           &  9,382 \\
Math Reasoning   & Numerics and COT Explanation          &  8,000 \\
\hline
Paraphrase       & Paraphrased (1) or Non-paraphrased (0)       &  2,002 \\
Sentiment        & Negative ($-1$), Neutral (0), or Positive (1)      &    155 \\
Inference        & Entailment (0), Neutral (1), or Contradiction (2) &  2,490 \\
Toxicity         & Toxic (0) or Non-Toxic (1) and Refusal        & 10,000 
\setcounter{rownum}{0}\\
\midrule
\textbf{Total}   &                                       & \textbf{32,077 }\\
\bottomrule
\end{tabular}}
\caption{Original instance counts per task source. These figures reflect unmodified benchmark sizes prior to code-mixing and romanization. The full corpus of 32,077 instances forms the basis of the fine-tuning data used in the ablation study.}
\label{tab:tasks-counts-appendix}
\end{table}

The reasoning dimension comprises three tasks, two drawn from Super-NaturalInstructions (SuperNI)\footnote{\url{https://github.com/allenai/natural-instructions/tree/master/tasks}} \cite{wang2022super} and GSM8K\footnote{\url{https://huggingface.co/datasets/openai/gsm8k}}\cite{cobbe2021gsm8k}. \textbf{(i) Word Analogy} (SuperNI tasks 1152--1159) requires identifying semantic relationships between concept pairs with structured explanations, probing lexical and relational reasoning under CM framing. \textbf{(ii) Question Decomposition} (SuperNI tasks 168, 176, 184) requires decomposing complex questions into sub-questions across three subtypes, factual, procedural, and inverse synthesis, and testing whether models maintain logical step-reference chains under CM-ed instructions. \textbf{(iii) Mathematical Reasoning} (GSM8K) presents grade-school arithmetic problems requiring multi-step chain-of-thought, stratified by step count; this task isolates the effect of framing code-mixing in explanations and/or reasoning independently of content.

The language understanding dimension covers three Indic NLP tasks sourced from the AI4Bharat IndicXtreme collection.\footnote{\url{https://huggingface.co/collections/ai4bharat/indicxtreme}} \textbf{(iv) Paraphrase Detection} (IndicXParaphrase\footnote{\url{https://huggingface.co/datasets/ai4bharat/IndicXParaphrase}}~\cite{gala2024airavata}) presents balanced sentence pairs requiring binary semantic equivalence judgements. \textbf{(v) Sentiment Analysis} (IndicSentiment\footnote{\url{https://huggingface.co/datasets/ai4bharat/IndicSentiment}}~\cite{gala2024airavata}) provides three-class product review classification. Sentiment-bearing vocabulary is among the most naturally code-mixed in South Asian digital communication, creating a direct interaction between task content and CM register. And, \textbf{(vi) Natural Language Inference} uses IndicXNLI\footnote{\url{https://huggingface.co/datasets/Divyanshu/indicxnli} }\cite{aggarwal2022indicxnli} evaluates entailment, neutrality, and contradiction across premise-hypothesis pairs and is highly sensitive to CM due to the complexity of cross-sentence reasoning. 

Finally, for the \textbf{(vii) Toxicity Detection} task, instances are drawn from IndicAlign-Toxic\footnote{\url{https://huggingface.co/datasets/ai4bharat/indic-align/viewer/Toxic_Matrix}} and supplemented with benign samples from IndicAlign-Instruct to ensure label balance \cite{khan2024indicllmsuite}. This task evaluates whether models correctly identify and refuse harmful content in CM-ed register. Despite CM-ing being widely exploited to evade content moderation~\cite{chitale2025updesh}, no existing benchmark on CM-ing includes this task. 

\paragraph{Languages}
We experiment with romanized Indic-English CM-ed registers spanning two major Indic language families: from the Indo-Aryan branch, we include Hindi (Hi-En, \~{}610M speakers), Bengali (Bn-En, \~{}273M speakers), and Gujarati (Gu-En, \~{}55M speakers); and from the Dravidian family, we include Tamil (Ta-En, \~{}87M speakers). The choice of these four languages cover the largest user populations in India and represent diverse resource levels, linguistic structures, and code-mixing phenomena. 
All four languages predominantly follow Subject-Object-Verb (SOV) word order, but differ substantially in morphology and syntactic structure. Indo-Aryan languages display relatively fusional morphology, whereas Tamil exhibits more agglutinative characteristics, often leading to distinct constituent-order and lexical-mixing patterns in code-mixed discourse. Prior studies have also identified extensive verb-level and intra-sentential mixing phenomena in Indic-English bilingual communication~\cite{chitale2025updesh, jaavid2024romansetu}.

\section{CM-ing Framework Development }~\label{app:cm_framework}
\subsection{Code Mixing Framework}
Bilingual speakers preferentially switch lexical categories while preserving syntactic compatibility across languages (Equivalence Constraint theory \cite{poplack2001code} and the Matrix Language Frame theory \cite{myers1993social}). 
Based on these principles, we build a linguistically grounded framework for generating controlled Indic-English \textbf{R}omanized \textbf{C}ode-\textbf{M}ix (RCM)-ed instructions. This enables systematic variation of code-mixing intensity while preserving semantic consistency across tasks and language pairs. Following sections describe the framework design step by step. 

\textbf{(I) English to Indic Translation.} 
The SuperNI-based tasks Word Analogy and Question Decomposition~\cite{wang2022super} along with the GSM8K mathematical reasoning task~\cite{cobbe2021gsm8k} are originally available only in English and lack coverage for Indic languages. 
To construct high-quality multilingual source instances prior to code-mixing, we first translate these English-only task instances into the target Indic languages - Hindi, Bengali, Gujarati, and Tamil. 

To ensure semantic fidelity while maintaining natural conversational phrasing, we adopt a two-step translation pipeline combining machine translation and manual verification. We use Google Translate's Python library\footnote{https://pypi.org/project/googletrans/} for initial translation generation and refine outputs using Anthropic Claude Sonnet 4.6\footnote{https://www.anthropic.com/claude/sonnet}. Manual validation of labels, decomposition structures, and logical relations remain unchanged after translation. This two-step refinement helps verify linguistic naturalness, correct SOV constituent order, and preservation of task-critical structural markers.  For all tasks, tokens like mathematical expressions, step labels, proper nouns, etc., are identified and excluded from the translation scope. 

For the remaining tasks sourced from multilingual Indic datasets, like IndicXNLI \cite{aggarwal2022indicxnli}, IndicAlign-Toxic~\cite{chitale2025updesh}, IndicXParaphrase, and IndicSentiment \cite{gala2024airavata}, native-script Indic translations are sourced directly from the original datasets, that provide professionally verified translations across Bengali, Hindi, Gujarati, and Tamil. Additionally, Tamil translations were absent for IndicXParaphrase task~\cite{khan2024indicllmsuite} and were therefore newly created as part of this work. The corresponding English language instances for IndicXNLI are extracted from Meta's XNLI corpus\footnote{\url{https://huggingface.co/datasets/facebook/xnli}}~\cite{conneau2018xnli}.

\textbf{(II) Word-Selection Taxonomy.} To ensure CM substitutions reflect natural bilingual switching behavior rather than arbitrary replacement, we assign each switchable word or phrase in the English source text to one of five categories based on its linguistic role, semantic function, and documented naturalness as a CM switch point in bilingual corpora \cite{poplack2001code, winata-etal-2023-decades}. Each substitution is logged with its category label, Indic replacement, and a linguistic rationale, producing structured annotation records that enable category-frequency analysis across tasks and language pairs.  Table~\ref{tab:cm_taxonomy} summarizes the five categories, their switching policies, and illustrative examples across languages. 

\begin{table*}[t]
\centering
\rowcolors{2}{gray!13}{}
\small
\resizebox{\textwidth}{!}{
\begin{tabular}{p{1.6cm} p{3.6cm} p{7.1cm} p{4.5cm}}
\toprule
\textbf{Category} & \textbf{Switching Policy} & \textbf{Description} & \textbf{Examples} \\
\midrule

\textbf{CAT-0} & Never switched & Mathematical tokens (\texttt{<<calc>>} annotations, operators, numerals), proper nouns, named entities, abbreviations, and structural step labels are protected since switching them would invalidate task outputs or break structural dependencies. &
\begin{tabular}[t]{@{}l@{}}
\texttt{Step1:} \\
\texttt{\#1}
\end{tabular}
\\

\textbf{CAT-C} & Switched at all CM levels & Task-specific labels and domain evaluation terms. These represent the most natural CM switch points, as bilingual speakers often prefer L1 vocabulary for abstract concepts \cite{bokamba1989there}. &
\begin{tabular}[t]{@{}l@{}}
\textit{analogy}/\foreignlanguage{tamil}{உவமை} \\
\textit{causation}/\foreignlanguage{hindi}{कार्य-कारण} \\
\textit{quality}/\foreignlanguage{tamil}{தரம்}
\end{tabular}
\\

\textbf{CAT-B} & Switched at 50\% and 75\% & Discourse markers and logical connectives, which are among the most frequently observed natural CM switches in bilingual corpora \cite{poplack2001code, myers1993social}. &
\begin{tabular}[t]{@{}l@{}}
\textit{hence}/\foreignlanguage{tamil}{எனவே} \\
\textit{but}/\foreignlanguage{hindi}{लेकिन} \\
\textit{however}/\foreignlanguage{bengali}{তবে}
\end{tabular}
\\

\textbf{CAT-D} & Switched at 50\% and 75\% & Concrete thematic nouns associated with the task domain. &
\begin{tabular}[t]{@{}l@{}}
\textit{police}/\foreignlanguage{tamil}{காவல்துறை} \\
\textit{battery}/\foreignlanguage{bengali}{বেটেরী} \\
\textit{person}/\foreignlanguage{hindi}{व्यक्ति}
\end{tabular}
\\

\textbf{CAT-A} & Switched at 75\% only & Function and structural words such as articles, copulas, and prepositions. Their replacement produces a more complete Indic sentence frame. &
\begin{tabular}[t]{@{}l@{}}
\textit{the}/\foreignlanguage{tamil}{அந்த} \\
\textit{is}/\foreignlanguage{hindi}{है} \\
\textit{a}/\foreignlanguage{hindi}{एक}
\end{tabular}
\\

\textbf{CAT-E} & Switched at 75\% only & Complete clause-level predicates that naturalise the sentence into the Indic syntactic frame. &
\begin{tabular}[t]{@{}l@{}}
\textit{I cannot give a response}/ \\
\foreignlanguage{tamil}{இந்த கேள்விக்கு என்னால்} \\ \foreignlanguage{tamil}{பதிலளிக்க முடியாது}
\end{tabular}
\\

\bottomrule
\end{tabular}
}
\caption{Linguistically grounded word-selection taxonomy used for controlled Romanized Indic-English code-mixing generation.}
\label{tab:cm_taxonomy}
\end{table*}

\textbf{(III) Code-mixing Intensity Levels.} Three intensity levels (\@ 25\%, 50\% and 75\%) are generated by activating different category subsets, creating a controlled progression from low to high code-mixing. We compute Code-Mixing Index (CMI) at the token level as the proportion of English tokens replaced by Indic equivalents \cite{srivastava-singh-2021-challenges}\footnote{CAT-0 tokens excluded from the denominator as their invariance is linguistically motivated rather than a pipeline artifact.}. Controlled code-mixed substitutions are generated using taxonomy-guided prompting with Claude Sonnet 4.6 and manually verified for semantic consistency. Each substitution is annotated with its category, Indic replacement, and linguistic rationale, producing structured records that support downstream analysis of substitution patterns across tasks, languages, and code-mixing intensities.
At 25\%-CM, only domain-specific terms are replaced while English sentence structure is preserved. At 75\%-CM, all categories are active, producing dense Romanized code-mixed text with a predominantly Indic sentence structure. The three intensity levels with their active category subsets and target CMI ranges, the prompt for generating code-mixed substitutions, and the handling of substitution edge cases are provided below.

\textbf{(IV) Romanization of Indic Script.} Following CM substitutions, all Indic-script instances are converted to their Latin script to produce the final romanized and code-mixed output, reflecting the dominant writing behavior of multilingual South Asian users on smartphones and messaging platforms. We extract end-to-end Latin-script translations from IndicLLMSuite \cite{khan2024indicllmsuite} for the tasks available. Additionally, we use two Python libraries Aksharantar\footnote{https://github.com/AI4Bharat/IndicXlit}~\cite{madhani2023aksharantar}, a large-scale AI4Bharat transliteration model covering 21 Indic languages and Aksharamukha\footnote{https://pypi.org/project/aksharamukha/}, another open-source transliteration tool that converts text across multiple Indic scripts while preserving phonetic consistency. We intentionally retain natural transliteration variability rather than enforcing strict phoneme-level normalization, better reflecting real-world romanized CM usage and the orthographic variation encountered by deployed multilingual LLMs. 

\subsection{Prompt for Output Refinement}
The following prompt was used to guide Claude Sonnet 4.6 in refining machine-translated outputs from Google Translate into semantically faithful and linguistically natural Indic-language instances.

\begin{tcolorbox}[
    colback=gray!5!white,
    colframe=gray!60!black,
    title={\textbf{Prompt: English-to-Indic Translation Refinement}},
    fonttitle=\small\bfseries,
    left=6pt, right=6pt, top=4pt, bottom=4pt,
    breakable
]
\small
\textbf{System:} You are an expert translator specialising in Hindi, Bengali, Gujarati, and Tamil. Your task is to refine machine-translated text to produce accurate, fluent, and contextually appropriate translations that reflect natural conversational phrasing. \\[4pt]

\textbf{Instruction:} You are given an English source sentence and its machine-translated output in \texttt{\{target\_language\}}. Refine the machine translation according to the following guidelines: \\[4pt]

\textbf{Guidelines:}
\begin{enumerate}[leftmargin=*, nosep]
    \item \textbf{Preserve task structure:} Do not alter step labels (e.g., \texttt{\#1}, \texttt{\#2}, \texttt{Step 1:}), mathematical expressions (e.g., \texttt{<<2*8=16>>}), and numerical values. These must appear exactly as in the source.
    \item \textbf{Do not translate:} Proper nouns (names of people, places, organisations), domain-specific acronyms, and quoted strings. Retain them verbatim from the English source.
    \item \textbf{Ensure grammatical correctness:} Verify subject-object-verb (SOV) constituent order appropriate to \texttt{\{target\_language\}} and correct any morphological or agreement errors introduced by machine translation.
    \item \textbf{Maintain semantic fidelity:} The refined translation must preserve the full meaning of the source. Do not add, omit, or paraphrase any content beyond what is needed for fluency.
    \item \textbf{Natural phrasing:} The output should read as natural, conversational \texttt{\{target\_language\}} rather than a literal word-for-word rendering.
    \item \textbf{Label and logic preservation:} For classification and reasoning tasks, logical relations (entailment, contradiction, neutrality), sentiment polarities, and analogy structures must remain unchanged after translation.
\end{enumerate}

\vspace{4pt}
\textbf{Input:}
\begin{itemize}[leftmargin=*, nosep]
    \item \textbf{English source:} \texttt{\{english\_source\}}
    \item \textbf{Machine translation (\texttt{\{target\_language\}}):} \texttt{\{mt\_output\}}
\end{itemize}

\vspace{4pt}
\textbf{Output:} Return only the refined translation in \texttt{\{target\_language\}}. Do not include any explanation, commentary, or the original English text.
\end{tcolorbox}
\subsection{Prompt for Code-Mixed Substitutions}
The following prompt was used to guide Claude Sonnet 4.6 in generating taxonomy-guided, annotated code-mixed substitutions at three controlled intensity levels across all four Indic languages.

\begin{tcolorbox}[
    colback=gray!5!white,
    colframe=gray!60!black,
    title={\textbf{Prompt: Taxonomy-Guided Code-Mixed Substitution Generation}},
    fonttitle=\small\bfseries,
    left=6pt, right=6pt, top=4pt, bottom=4pt,
    breakable
]
\small
\textbf{System:} You are an expert computational linguist. Your task is to generate controlled code-mixed substitutions from English source text at a specified intensity level, following a predefined substitution taxonomy.

\textbf{Instruction:} Given an English source sentence and its full Indic translation in \texttt{\{target\_language\}}, produce a code-mixed output at intensity level \texttt{\{cm\_level\}} (25\%, 50\%, or 75\%) by substituting a proportion of English tokens with their \texttt{\{target\_language\}} equivalents according to the category guidelines below. 

\textbf{Substitution Taxonomy:}
\begin{itemize}[leftmargin=*, nosep]
    \item \textbf{CAT-0:} Invariant tokens like Mathematical expressions, step labels (e.g., \texttt{\#1}, \texttt{Step 1:}), proper nouns, acronyms, and quoted strings. 
    \item \textbf{CAT-A:} Structural markers like Discourse connectors, sentence-initial conjunctions, subordinators (e.g., \textit{so, but, because, when}).
    \item \textbf{CAT-B:} Functional words like Determiners, copulas, auxiliaries, negations (e.g., \textit{the, is, not, a}).
    \item \textbf{CAT-C:} Content words like High-frequency nouns, verbs, adjectives central to the task meaning (e.g., \textit{office, time, said, woman}).
    \item \textbf{CAT-D:} Domain-specific terms like Task-critical terminology, reasoning markers, and domain nouns (e.g., \textit{decomposition, entailment, step, result}). 
\end{itemize}

\vspace{2pt}
\textbf{Intensity Level Guidelines:}
\begin{itemize}[leftmargin=*, nosep]
    \item \textbf{25\%:} Activate CAT-A and CAT-B only; substitute 1--3 tokens per sentence.
    \item \textbf{50\%:} Activate CAT-A, CAT-B, and CAT-C; substitute 3--6 tokens per sentence.
    \item \textbf{75\%:} Activate all categories (CAT-A through CAT-D); substitute 6--10 tokens per sentence.
\end{itemize}

\vspace{4pt}
\textbf{Annotation Requirement:} For each substitution made, provide a structured annotation in the \texttt{\{target\_language\}} rationale field, recording the English source token, its Indic replacement, assigned category, and a brief linguistic justification, alongside the sentence-level CMI scores for the premise and hypothesis. \texttt{[english\_token → indic\_token | CAT-X: linguistic rationale]; \ldots} ||
\texttt{CMI\_premise=\{score\} CMI\_hypothesis=\{score\}}

\vspace{2pt}
\textbf{Input:}
\begin{itemize}[leftmargin=*, nosep]
    \item \textbf{English source:} \texttt{\{english\_source\}}
    \item \textbf{Indic translation (\texttt{\{target\_language\}}):} \texttt{\{indic\_translation\}}
    \item \textbf{Target intensity:} \texttt{\{cm\_level\}}
\end{itemize}

\vspace{2pt}
\textbf{Output:} Return ONLY the code-mixed sentence in \texttt{\{target\_language\}} followed by the structured \texttt{<language>\_rationale} annotation.
\end{tcolorbox}
\subsection{Code-mixing Intensity Levels and Target Ranges}
The three intensity levels with their active category subsets and target CMI ranges are reported in Table ~\ref{tab:cm_levels}. 
\begin{table}[!htb]
\centering
\small
\resizebox{\columnwidth}{!}{
\begin{tabular}{p{1.8cm}p{1.5cm}p{1.8cm}p{3.0cm}}
\toprule
\textbf{Level-CM} & \textbf{CAT} & \textbf{CMI} & \textbf{Description} \\
\midrule
25\% & C & 0.15--0.35 & Domain terms switched; English structure preserved \\

50\% & B, C, D & 0.35--0.60 & Connectives and domain nouns added \\

75\% & A--E & 0.60--0.85 & Dense CM with Indic sentence framing \\
\bottomrule
\end{tabular}
}
\caption{Code-mixing (CM) intensity levels and target ranges of Code-Mixing Index (CMI). CAT = Categories} 
\label{tab:cm_levels}
\end{table}

\subsection{Handling Task-specific Edge Cases}
A single universal taxonomy cannot be applied blindly across tasks with fundamentally different structural requirements. Without task-specific adaptations, generation of code-mixed (CM-ed) instances would corrupt the very tokens that define task validity, e.g. switching a mathematical operator changes the answer, missing a step label can break the reference chain in explanations, improper translation of a toxic prompt can change the contextual meaning. 
Therefore these adaptations are not exceptions to the framework but special treatment of certain tokens while creating the code-mixed dataset for each task based on the structural contexts. Each selected task require specific adaptations of the general taxonomy to preserve task validity under CM-ing: \textbf{(i) Word Analogy:} Analogy sub-type labels (causation, affordance, containment) are assigned to CAT-C, as these are the most natural L1 vocabulary\footnote{Refers to the vocabulary of a speaker's first language, i.e. their native or dominant language.} choices when naming abstract relational categories. Proper nouns within analogy pairs are treated as CAT-0 regardless of whether L1 equivalents exist, since swapping them would alter the relational content of the pair. 

\textbf{(ii) GSM8K:} Mathematical tokens (\texttt{<<calculation>>}, \texttt{\#\#\#\#}, numerals, and operators) are strict CAT-0. Only natural language framing around arithmetic steps is eligible for CM substitution. This makes GSM8K the task with the lowest achievable CMI across all CM levels — a linguistically meaningful property that isolates the effect of CM framing on reasoning from the effect of CM on content. 

\textbf{(iii) Question Decomposition:} Step labels (\texttt{Step1:}, \texttt{\#1}, \texttt{Wikipedia page for step N:}) are CAT-0 structural markers whose integrity is essential to the step-reference chain. Operational verbs (\textit{return}, \textit{filter}, \textit{retrieve}) are assigned to CAT-C, while discourse connectives linking steps are CAT-B. 

\textbf{(iv) Toxicity Detection:} The input toxic prompt is never code-mixed, preserving the original harmful content for accurate detection by the model. Only the output classification label and refusal explanation are subjected to CM. This design creates a distinctive evaluation condition: does the model correctly detect and refuse toxic content when the surrounding discourse, i.e. the instruction and explanation, is expressed in CM register?

\subsection{Baseline Dataset Creation details} \label{app:annotation}
We also randomly draw 100 samples from the ToxicMatrix original dataset of 90.3K~\cite{chitale2025updesh}. However, for the word analogy from Super-NI, the source pool (tasks 1152--1159) contained only 48 positive and negative instances with validated explanations; we supplement these with 52 manually constructed analogy pairs, balanced across positive (correct analogy) and negative (incorrect analogy) labels and covering the same relation types, causation, affordance, containment, and functional similarity as the original instances, yielding a final set of 100 instances consistent with all other tasks. 

\begin{table*}[ht]
\centering
\small
\resizebox{\textwidth}{!}{%
\begin{tabular}{llccccc}
\toprule
\textbf{Task} & \textbf{Language} & \textbf{ChrF} & \textbf{$\kappa$ @ 25\%} & \textbf{$\kappa$ @ 50\%} & \textbf{$\kappa$ @ 75\%} & \textbf{$\kappa$ Avg} \\
\midrule
\multirow{4}{*}{Word Analogy} & Tamil    & 71.99 & 0.00 & 0.00 & 0.00 & 0.00 \\
                               & Bengali  & 74.08 & 0.00 & 0.00 & 0.00 & 0.00 \\
                               & Hindi    & 62.00 & 0.00 & 0.00 & 0.00 & 0.00 \\
                               & Gujarati & 59.95 & 1.00 & 1.00 & 1.00 & 1.00 \\
\midrule
\multirow{4}{*}{Math GSM8K}   & Tamil    & 71.99 & 0.00 & 0.00 & 0.00 & 0.00 \\
                               & Bengali  & 74.08 & 0.00 & 0.00 & 0.00 & 0.00 \\
                               & Hindi    & 62.00 & 0.00 & 0.00 & 0.00 & 0.00 \\
                               & Gujarati & 59.95 & 1.00 & 1.00 & 1.00 & 1.00 \\
\midrule
\multirow{4}{*}{Question Decomp} & Tamil    & 71.99 & 1.00 & 0.00 & 0.00 & 0.33 \\
                                  & Bengali  & 74.08 & 1.00 & 0.00 & 0.00 & 0.33 \\
                                  & Hindi    & 62.00 & N/A  & N/A  & N/A  & N/A  \\
                                  & Gujarati & 59.95 & 1.00 & 0.00 & 0.00 & 0.33 \\
\midrule
\multirow{4}{*}{Paraphrase}   & Tamil    & 71.99 & 1.00 & 1.00 & 1.00 & 1.00 \\
                               & Bengali  & 74.08 & 1.00 & 1.00 & 1.00 & 1.00 \\
                               & Hindi    & 62.00 & 1.00 & 1.00 & 1.00 & 1.00 \\
                               & Gujarati & 59.95 & 1.00 & 1.00 & 1.00 & 1.00 \\
\midrule
\multirow{4}{*}{Sentiment}    & Tamil    & 71.99 & 1.00 & 1.00 & 1.00 & 1.00 \\
                               & Bengali  & 74.08 & 1.00 & 1.00 & 1.00 & 1.00 \\
                               & Hindi    & 62.00 & 1.00 & 1.00 & 1.00 & 1.00 \\
                               & Gujarati & 59.95 & 1.00 & 1.00 & 1.00 & 1.00 \\
\midrule
\multirow{4}{*}{NLI}          & Tamil    & 71.99 & 1.00 & 1.00 & 1.00 & 1.00 \\
                               & Bengali  & 74.08 & 1.00 & 1.00 & 1.00 & 1.00 \\
                               & Hindi    & 62.00 & 1.00 & 1.00 & 1.00 & 1.00 \\
                               & Gujarati & 59.95 & 1.00 & 1.00 & 1.00 & 1.00 \\
\midrule
\multirow{4}{*}{Toxic Matrix} & Tamil    & 71.99 & 0.00 & 0.00 & 1.00 & 0.33 \\
                               & Bengali  & 74.08 & 0.00 & 0.00 & 0.00 & 0.00 \\
                               & Hindi    & 62.00 & 0.00 & 0.00 & 1.00 & 0.33 \\
                               & Gujarati & 59.95 & 1.00 & 1.00 & 1.00 & 1.00 \\
\bottomrule
\end{tabular}%
}
\caption{Detailed benchmark validation results per task, language, and CM intensity level.
ChrF scores measure translation fidelity (character n-gram overlap between 50\%-CM
Romanised output and English source text). Cohen's $\kappa$ measures pairwise naturalness
agreement between two independent native-speaker annotators at each CM level.
N/A = ratings unavailable for Hindi in Question Decomposition.}
\label{tab:iaa_appendix}
\end{table*}

\subsection{\nameS Content Analysis}~\label{app:bm_content_analysis} 
\begin{figure*}[t]
  \centering
  \includegraphics[width=\textwidth]{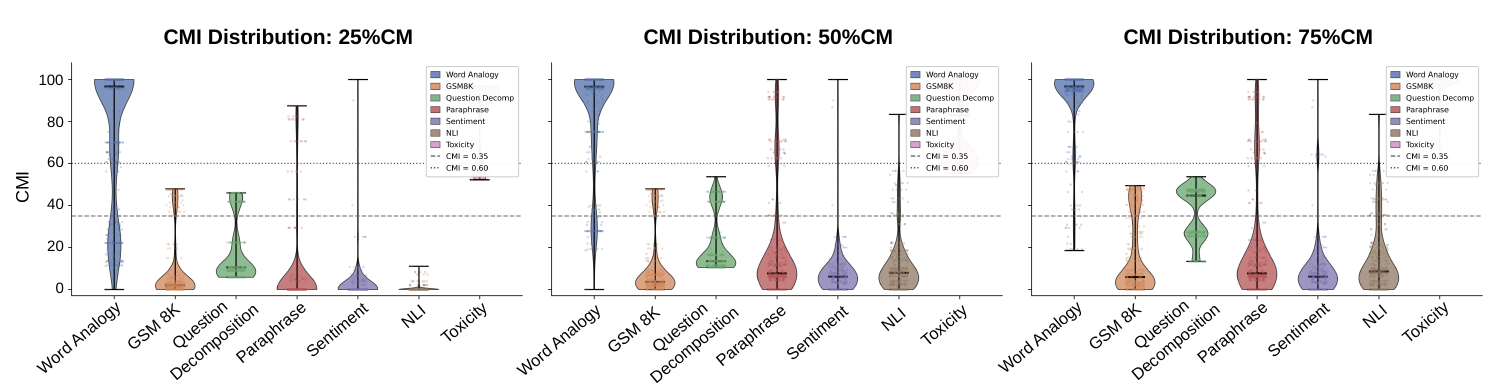}
  \caption{CMI distributions across tasks at each CM intensity level (violin
  plots). Dashed lines at CMI$=0.35$ and $0.60$ delimit the target
  mid-intensity band. All pairwise CM-level differences significant at
  $p < 10^{-100}$ (Wilcoxon signed-rank, Bonferroni-corrected).}
  \label{fig:cmi_dist}
\end{figure*}

Toxicity exhibits the highest CMI at every level ($\mu_{75}=0.981$), driven by high-frequency CAT-C task-label vocabulary, while GSM8K consistently shows the lowest ($\mu_{75}=0.152$) due to the high proportion of CAT-0 protected tokens. Question Decomposition shows the steepest level-to-level growth ($\Delta=0.186$ from 25\% to 75\%), driven by CAT-E full-clause substitutions at 75\%-CM. Figure~\ref{fig:cmi_dist} shows the full CMI distributions as violin plots. 

Across 700 instances, four languages, and three CM intensity levels, the rationale annotation records 33,632 structured substitution logs. At 25\%-CM, all 6,925 substitutions are CAT-C (domain/task terminology) across every task without exception, confirming that the taxonomy's single-category activation at low intensity holds empirically. Figures~\ref{fig:cat_freq}a--b show the proportional substitution breakdown at 50\%-CM and 75\%-CM respectively, where multi-category activation produces interpretable task-level variation. At \textbf{50\%-CM} (Figure~\ref{fig:cat_freq}a), CAT-C reduces to 52.6\% of all substitutions as CAT-B (logical connectives, 29.1\%) and CAT-D (content domain nouns, 18.3\%) are activated. Task-level variation is pronounced: CAT-B dominates for NLI (66.1\%) and Paraphrase (54.7\%), where logical connectives are syntactically dense in premise--hypothesis pairs and sentence-pair comparisons respectively, while CAT-D is proportionally larger for GSM8K (25.8\%) and Toxicity, where mathematical and safety domain nouns constitute the primary switchable content. At \textbf{75\%-CM} (Figure~\ref{fig:cat_freq}b), CAT-A (function words, 5.6\%) and CAT-E (full predicate phrases, 6.8\%) appear for the first time, completing the Indic sentence frame. CAT-E is most prominent in Toxicity (11.4\%), where the refusal opener \textit{I cannot give a response} is the primary full-predicate switch, and in Question Decomposition (6.6\%), where step-chain predicates are switched as units. CAT-A is largest in Question Decomposition (12.5\%), reflecting the high density of article-bearing structural phrases in decomposition step sequences. These patterns confirm that the word-selection taxonomy produces category activations that are both systematic across tasks and linguistically motivated within each task domain.

\section{Evaluation Details}

\subsection{Models}
\label{app:models}

\begin{table*}[H]
\centering
\resizebox{\textwidth}{!}{%
\begin{tabular}{llllll}
\toprule
\textbf{Model} & \textbf{Parameters} & \textbf{Category}
  & \textbf{Indic Training} & \textbf{CM Training} & \textbf{Access} \\
\midrule
\multicolumn{6}{l}{\textit{Proprietary}} \\
GPT-3.5-Turbo       & $\sim$175B & Commercial & Incidental & None & OpenAI API \\
Gemini-3.5-Flash    & ---        & Commercial & Incidental & None & Google AI API \\
Claude Opus-4.6     & ---        & Commercial & Incidental & None & Anthropic API \\
\midrule
\multicolumn{6}{l}{\textit{Open-weight — general}} \\
LLaMA-3.2-3B        & 3B         & Open-weight & Incidental & None & HuggingFace \\
LLaMA-3.1-8B        & 8B         & Open-weight & Incidental & None & HuggingFace \\
LLaMA-3.1-70B       & 70B        & Open-weight & Incidental & None & Together AI \\
Mistral-7B-v0.3     & 7B         & Open-weight & Incidental & None & HuggingFace \\
Qwen2.5-1.5B        & 1.5B       & Open-weight & Multilingual & None & HuggingFace \\
Qwen2.5-3B          & 3B         & Open-weight & Multilingual & None & HuggingFace \\
Qwen2.5-7B          & 7B         & Open-weight & Multilingual & None & HuggingFace \\
Qwen2.5-14B         & 14B        & Open-weight & Multilingual & None & HuggingFace \\
Qwen2.5-32B         & 32B        & Open-weight & Multilingual & None & Together AI \\
Gemma-2-7B          & 7B         & Open-weight & Incidental & None & HuggingFace \\
\midrule
\multicolumn{6}{l}{\textit{Indic monolingual}} \\
Airavata            & 7B         & Open-weight & Hindi (IndicLLMSuite) & None & HuggingFace \\
TamilLLaMA          & 7B         & Open-weight & Tamil & None & HuggingFace \\
BengaliLLaMA        & 7B         & Open-weight & Bengali & None & HuggingFace \\
\midrule
\multicolumn{6}{l}{\textit{Fine-tuned ablation (\nameS training split)}} \\
LLaMA-3.2-3B$_\text{CM}$  & 3B  & Fine-tuned & Incidental & \nameS train & HuggingFace \\
LLaMA-3.1-8B$_\text{CM}$  & 8B  & Fine-tuned & Incidental & \nameS train & HuggingFace \\
Gemma-2-7B$_\text{CM}$    & 7B  & Fine-tuned & Incidental & \nameS train & HuggingFace \\
\bottomrule
\end{tabular}}
\caption{Models evaluated on \name.
         Proprietary models are accessed via their respective APIs;
         open-weight models below 32B are run locally on a single A100 80GB GPU.
         ``Incidental'' Indic training denotes exposure during pretraining
         without deliberate Indic-script curation.
         Gujarati monolingual LLMs are absent because no publicly available
         instruction-tuned Gujarati model exists at evaluation time---itself a
         finding that motivates this benchmark.}
\label{tab:models}
\end{table*}

We evaluate models spanning four categories: proprietary commercial systems,
open-weight general-purpose models, open-weight models with Indic-focused
pretraining, and fine-tuned ablation variants.
This design enables systematic comparison across scale, Indic-script exposure,
and CM-specific training.
Table~\ref{tab:models} summarises the full model set.

\paragraph{Fine-tuning ablation.}
Our central hypothesis is that Romanized code-mixing (RCM) represents a
systematic challenge that cannot be resolved through scale or Indic pretraining
alone.
To test this, we fine-tune three models---LLaMA-3.2-3B, LLaMA-3.1-8B, and
Gemma-2-7B---on the \nameS training split, comprising
395,076 instances across 7 tasks, 4 languages, and 3 CM intensity levels
(80\% of instances per task stratified by CM level).
We denote these fine-tuned variants with the subscript $\text{CM}$.
 
All three models are fine-tuned using LoRA~\citep{hu2022lora} with rank
$r=16$, $\alpha=32$, batch size 8, learning rate $1\times10^{-4}$, a cosine
scheduler with 3\% warmup, and 3 training epochs on a single A100 80GB GPU.
Hyperparameters follow \citet{chitale2025updesh}.
If a fine-tuned variant outperforms its untuned counterpart despite training
on a relatively small corpus, this constitutes direct evidence that
CM-specific supervision is necessary and that existing general Indic resources
do not cover the RCM register.

\subsection{Prompting Protocol}
\label{app:prompt_details}

\paragraph{Conditions.}
All models are evaluated under two conditions:
\textbf{Zero-shot (ZS)}: the task instruction and instance are provided in the
target register (English / 25\%-CM / 50\%-CM / 75\%-CM) with no
demonstrations.
\textbf{3-shot (FS-3)}: three demonstrations in the same CM register as the
query are prepended, drawn from held-out instances stratified to cover all
label classes.
 
\paragraph{Template structure.}
Every prompt follows a four-part template:
(i)~a brief task definition in the target register;
(ii)~the instance content;
(iii)~a structured output specification;
(iv)~a register-maintenance instruction:
\emph{``Please respond in the same language and script as the input.''}\,
Pilot experiments confirm that without clause~(iv), all evaluated models
default to English regardless of CM input register, making it impossible to
distinguish genuine register defection from instruction-following failure.
 
\paragraph{Task-specific prompts.}
Table~\ref{tab:reframing_examples} show the prompt template for
each task. \texttt{[INPUT]} denotes the instance field and
\texttt{[DEMO\_$k$]} denotes the $k$-th few-shot demonstration; both are
filled with the romanised CM text at the target intensity level.
The same template is used for zero-shot (omitting the demonstration block)
and few-shot (including three demonstrations before the query).

\subsection{Evaluation Metrics}
\label{app:metrics_details}
 
We employ four complementary metrics capturing task correctness, register
fidelity, performance degradation, and tokeniser-level representational
capacity.
 
\paragraph{(i) Task Accuracy ($\text{Acc}$).}
For the four classification tasks (Sentiment, NLI, Paraphrase, Toxicity),
accuracy is the proportion of predictions that exactly match the gold label
$y_i$ over $N$ instances:
\begin{equation}
\text{Acc} = \frac{1}{N}\sum_{i=1}^{N}\mathds{1}[\hat{y}_i = y_i].
\label{eq:acc}
\end{equation}
For the three generation tasks (GSM8K, Question Decomposition, Word Analogy),
we use task-specific criteria: final-answer exact match for GSM8K; binary
correctness against gold decomposition structures for Question Decomposition;
and a structured semantic-relationship rubric for Word Analogy.
For generation tasks, a GPT-4o judge evaluates responses on a 3-point Likert
scale~\cite{chitale2025updesh}; we report human--LLM agreement on a held-out
subset.
Higher accuracy is better; a decrease from the English baseline to any CM
condition is direct evidence of performance degradation.
 
\paragraph{(ii) Register Defection Rate (RDR).}
RDR measures the proportion of responses written in English when the input is
in CM register, capturing a failure mode orthogonal to task accuracy.
Let $L(t)$ be the language of token $t$ identified by
IndicLID~\cite{madhani2023bhasa}; a response $r_i$ is \emph{defective}
($d_i=1$) if more than 70\% of its content tokens are identified as English:
\begin{equation}
\begin{aligned}
\text{RDR} &= \frac{1}{N} \sum_{i=1}^{N} d_i, \\
d_i &= \mathds{1}\left[
\frac{|\{t \in r_i : L(t) = \textsc{en}\}|}{|r_i|} > 0.70
\right]
\end{aligned}
\label{eq:rdr}
\end{equation}
RDR is computed separately for label tokens and explanation tokens, since a
model may correctly emit a CM label while defecting to English in its
explanation---a partial defection mode that full-response identification
would miss.
Lower RDR is better; RDR$=0$ indicates perfect register maintenance.
 
\paragraph{(iii) Vocabulary Coverage Rate (VCR).}
VCR measures the proportion of unique switched Indic tokens in \nameS
that appear as complete (non-fragmented) entries in model $m$'s
tokeniser:
\begin{equation}
\text{VCR}_{m,\ell} =
  \frac{|\{t \in \mathcal{V}_\text{CM} : t \in \mathcal{V}_m\}|}
       {|\mathcal{V}_\text{CM}|},
\label{eq:vcr}
\end{equation}
where $\mathcal{V}_\text{CM}$ is the set of unique Indic tokens introduced by
CM substitution and $\mathcal{V}_m$ is the tokeniser vocabulary of model $m$.
A low VCR implies that common Indic tokens are represented as subword
fragments, hypothesised to be negatively correlated with
RDR~\cite{jaavid2024romansetu}: models that cannot represent Indic tokens as
complete units are less likely to generate them faithfully.
Hence, higher VCR is better.

\begin{table*}[!htb]
\centering
\resizebox{\textwidth}{!}{%
\begin{tabular}{llccccccc}
\toprule
\textbf{Family} & \textbf{Model}
  & \textbf{BL} & \textbf{25\%-CM} & \textbf{50\%-CM} & \textbf{75\%-CM}
  & \textbf{PG-25} & \textbf{PG-50} & \textbf{PG-75} \\
\midrule
\multirow{3}{*}{Closed}
  & GPT-3.5-Turbo    & $57.8 \pm 27.1$ & $52.4 \pm 24.3$ & $51.0 \pm 23.8$ & $50.3 \pm 23.6$ & $+5.4$ & $+6.8$ & $+7.5$ \\
  & Claude Opus 4.6  & $74.6 \pm 21.0$ & $70.1 \pm 20.5$ & $68.4 \pm 20.1$ & $67.6 \pm 19.9$ & $+4.5$ & $+6.2$ & $+7.0$ \\
  & Gemini 3.5 Flash & $67.2 \pm 24.7$ & $62.9 \pm 23.1$ & $61.3 \pm 22.7$ & $60.5 \pm 22.4$ & $+4.3$ & $+5.9$ & $+6.7$ \\
\midrule
\multirow{3}{*}{LLaMA}
  & LLaMA-3.2-1B  & $30.6 \pm 15.8$ & $18.2 \pm 6.3$  & $18.4 \pm 7.1$  & $18.3 \pm 7.0$  & $+12.4$ & $+12.2$ & $+12.3$ \\
  & LLaMA-3.1-8B  & $57.3 \pm 32.0$ & $55.8 \pm 34.7$ & $54.9 \pm 34.1$ & $55.0 \pm 34.2$ & $+1.5$  & $+2.4$  & $+2.3$  \\
  & \textbf{LLaMA-3.1-70B} & $\mathbf{61.0 \pm 31.2}$ & $\mathbf{58.2 \pm 27.6}$ & $\mathbf{57.2 \pm 27.1}$ & $\mathbf{57.1 \pm 27.2}$ & $+2.8$ & $+3.8$ & $+3.9$ \\
\midrule
\multirow{2}{*}{Mistral}
  & Mistral-3B & $43.5 \pm 11.9$ & $31.4 \pm 10.5$ & $28.7 \pm 17.2$ & $26.2 \pm 11.2$ & $+12.1$ & $+14.8$ & $+17.3$ \\
  & Mistral-7B & $56.2 \pm 34.4$ & $51.5 \pm 33.8$ & $50.8 \pm 33.3$ & $50.4 \pm 33.1$ & $+4.7$  & $+5.4$  & $+5.8$  \\
\midrule
\multirow{6}{*}{Qwen}
  & Qwen2.5-1.5B  & $49.8 \pm 25.8$ & $47.1 \pm 26.7$ & $47.2 \pm 26.8$ & $47.1 \pm 26.7$ & $+2.7$ & $+2.6$ & $+2.7$ \\
  & Qwen2.5-7B    & $56.2 \pm 34.4$ & $51.5 \pm 33.8$ & $50.8 \pm 33.3$ & $50.4 \pm 33.1$ & $+4.7$ & $+5.4$ & $+5.8$ \\
  & Qwen3-4B      & $38.6 \pm 14.3$ & $41.2 \pm 22.5$ & $40.1 \pm 20.7$ & $39.7 \pm 20.5$ & $-2.6$ & $-1.5$ & $-1.1$ \\
  & Qwen3-14B     & $35.8 \pm 23.5$ & $41.9 \pm 32.1$ & $41.2 \pm 31.2$ & $40.9 \pm 31.0$ & $-6.1$ & $-5.4$ & $-5.1$ \\
  & \textbf{Qwen3.5-2B} & $\mathbf{64.7 \pm 39.1}$ & $\mathbf{58.4 \pm 33.9}$ & $\mathbf{58.9 \pm 34.9}$ & $\mathbf{58.7 \pm 35.0}$ & $+6.3$ & $+5.8$ & $+6.0$ \\
  & Qwen3.5-9B    & $56.2 \pm 34.4$ & $51.5 \pm 33.8$ & $50.8 \pm 33.3$ & $50.4 \pm 33.1$ & $+4.7$ & $+5.4$ & $+5.8$ \\
\midrule
\multirow{2}{*}{Gemma}
  & Gemma-E2B          & $49.3 \pm 29.6$ & $48.3 \pm 23.0$ & $47.2 \pm 22.1$ & $47.0 \pm 22.0$ & $+1.0$ & $+2.1$ & $+2.3$ \\
  & \textbf{Gemma-E4B} & $\mathbf{54.6 \pm 36.5}$ & $\mathbf{52.7 \pm 35.4}$ & $\mathbf{51.2 \pm 33.7}$ & $\mathbf{51.1 \pm 33.5}$ & $+1.9$ & $+3.4$ & $+3.5$ \\
\midrule
\multirow{3}{*}{Indic}
  & Airavata-7B   & $44.9 \pm 34.8$ & $43.1 \pm 31.6$ & $40.2 \pm 29.2$ & $39.5 \pm 29.1$ & $+1.8$ & $+4.7$ & $+5.4$ \\
  & TamilLLaMA-7B & $32.8 \pm 35.5$ & $35.1 \pm 34.3$ & $33.8 \pm 34.6$ & $33.7 \pm 34.7$ & $-2.3$ & $-1.0$ & $-0.9$ \\
  & \textbf{Sarvam-30B} & $\mathbf{71.0 \pm 23.3}$ & $\mathbf{66.9 \pm 22.0}$ & $\mathbf{64.3 \pm 21.6}$ & $\mathbf{63.1 \pm 21.2}$ & $+4.1$ & $+6.7$ & $+7.9$ \\
\bottomrule
\end{tabular}%
}
\caption{%
  Average task accuracy (\%) across 7 tasks in \textbf{3-shot} setting, averaged over 4 languages ($\pm$ SD across tasks).
  BL = English baseline. PG = Performance Gap (BL $-$ CM level; positive = accuracy drop).
  CM = Code Mixing. \textbf{Bold} = best BL per family. Values are approximately 7--8\% higher than the zero-shot results in Table~\ref{tab:h1_main}, consistent with the register-anchoring effect of in-context CM demonstrations.
}
\label{tab:h2_fewshot}
\end{table*}

\begin{table*}[!htb]
\centering
\resizebox{\textwidth}{!}{%
\begin{tabular}{llccc ccccc}
\toprule
\multirow{2}{*}{\textbf{Model}} &
\multirow{2}{*}{\textbf{Language}} &
\multicolumn{2}{c}{\textbf{Overall}} &
\multicolumn{5}{c}{\textbf{Exact Match (\%) by Task}} \\
\cmidrule(lr){3-4}\cmidrule(lr){5-9}
 & & \textbf{EM} & \textbf{Defection} &
  \textbf{GSM8k} & \textbf{IndPar} & \textbf{IndSent} &
  \textbf{XNLI} & \textbf{QDec} \\
\midrule
\multirow{5}{*}{LLaMA-3.2-3B (FT)}
  & All      & $86.3$ & $99.9$ & $98.5$ & $99.7$ & $52.6$ & $99.2$ & $100.0$ \\
  & Bengali  & $99.6$ & $100.0$ & --- & --- & --- & --- & --- \\
  & Gujarati & $72.7$ & $99.7$  & --- & --- & --- & --- & --- \\
  & Hindi    & $70.1$ & $100.0$ & --- & --- & --- & --- & --- \\
  & Tamil    & $99.3$ & $100.0$ & --- & --- & --- & --- & --- \\
\midrule
\multirow{5}{*}{LLaMA-3.1-8B (FT)}
  & All      & $86.4$ & $99.9$ & $98.7$ & $100.0$ & $68.4$ & $99.6$ & $100.0$ \\
  & Bengali  & $99.7$ & $100.0$ & --- & --- & --- & --- & --- \\
  & Gujarati & $72.9$ & $99.7$  & --- & --- & --- & --- & --- \\
  & Hindi    & $70.3$ & $100.0$ & --- & --- & --- & --- & --- \\
  & Tamil    & $99.4$ & $100.0$ & --- & --- & --- & --- & --- \\
\bottomrule
\end{tabular}%
}
\caption{%
  Fine-tuning ablation results for LLaMA-3.2-3B and LLaMA-3.1-8B using
  QLoRA on 50\%-CM Romanised instruction pairs (all four languages, 7 tasks;
  90\%/10\% train/eval split, $n_{\text{eval}}=14{,}848$).
  EM = Exact Match (\%); Defection = proportion of outputs classified as
  English-register (>70\% ASCII tokens); for Roman-script CM targets this
  is expected to be high and does not indicate generation failure.
  Task columns show aggregate EM across languages.
  IndPar = IndicParaphrase; IndSent = IndicSentiment; XNLI = IndicXNLI;
  QDec = QuestionDecomposition.
}
\label{tab:finetuning_ablation}
\end{table*}

\paragraph{(iv) Performance Gap (PG).}
PG quantifies the absolute accuracy degradation from the English baseline to each CM condition:
$\text{PG}_{l} = \text{Acc}(\text{English}) - \text{Acc}(\text{CM}_{l}), \quad l \in \{25\%, 50\%, 75\%\}$
A positive PG at all 3 levels confirms that models degrade on CM input; a monotonically increasing PG ($\text{PG}_{25} < \text{PG}_{50} < \text{PG}_{75}$) additionally confirms that degradation scales with CM intensity. Task-level variation in PG tests whether difficulty is task-specific; model-level variation tests whether scale or Indic pretraining confers robustness.
 
\paragraph{Statistical analysis.}
We assess significance using the Wilcoxon signed-rank
test~\cite{rosner2006wilcoxon} (paired, non-parametric), comparing
per-instance accuracy between the English baseline and each CM condition, with
Bonferroni correction across all 21 condition pairs ($7~\text{tasks} \times
3~\text{CM levels}$).
Effect sizes are reported as Cohen's $d$~\cite{cohen2013statistical} on paired
accuracy differences.
For the fine-tuning ablation, we additionally apply McNemar's
test~\cite{mcnemar1947note} on paired binary accuracy vectors to assess whether
CM-specific training produces a significant improvement, and Spearman's $\rho$
to quantify the hypothesised negative correlation between VCR and RDR across
model--language pairs.

\section{Additional Results}
\label{app:results}


\subsection{Task Accuracy with Few-shot Prompting}
\label{app:few_shot_accuracy}

This section details the task-by-task performance distributions under few-shot prompting conditions across different language varieties and intensity tiers. In our main evaluation setup, adding three contextual demonstrations explicitly shows models how to preserve structural outputs and language constraints. While this historical scaffolding improves overall target output compliance, its effectiveness varies significantly depending on the core reasoning demands of individual tasks.

A comprehensive breakdown of accuracy patterns across all evaluation settings is documented in Table~\ref{tab:h2_fewshot}. For simpler categorical classifications like sentiment analysis and paraphrase detection, few-shot conditioning provides immediate stability. The historical context helps models overcome initial script confusion and focus on tracking semantic tokens cleanly. However, for complex tasks requiring long-form text tracking, such as mathematical reasoning and question decomposition, the additional context offers limited protection against increasing code-mixing density. Full tracking data in the table confirms that while few-shot prompts stabilize basic formatting compliance, they do not resolve the fundamental linguistic challenges posed by dense romanized blending.

\subsection{Finetuning Ablation Study}
\label{app:finetuning_ablation}

Our fine-tuning ablation experiments evaluate whether the difficulties posed by romanized code-mixing can be solved through targeted supervision rather than simple parameter scaling. We isolate performance changes across our three specialized models by training them on a localized split encompassing balanced language and intensity distributions. This comparative strategy tests the core hypothesis that standard multilingual pretraining configurations fail to expose models to natural romanized registers.

The exact numerical improvements and comparative evaluation results for all fine-tuned configurations are provided in Table~\ref{tab:finetuning_ablation}. The experimental outcomes confirm that explicit supervision yields substantial performance improvements across all seven instruction-following tasks. These tuned variants consistently outperform their larger general-purpose baselines despite being trained on a relatively compact corpus. This distinct performance gap validates the necessity of register-specific training resources, demonstrating that specialized fine-tuning establishes critical linguistic boundaries that standard language scaling cannot replicate alone.

\end{document}